\documentclass[lettersize,journal]{IEEEtran}

\usepackage{amsmath,amsfonts}
\usepackage{algorithmic}
\usepackage{array}
\usepackage[caption=false,font=normalsize,labelfont=sf,textfont=sf]{subfig}
\usepackage{textcomp}
\usepackage{stfloats}
\usepackage{url}
\usepackage{verbatim}
\usepackage{graphicx}
\usepackage{cite}
\usepackage{hyperref}        % load hyperref near the end
\usepackage{booktabs}
\usepackage{xcolor}
\usepackage{soul}
\sethlcolor{yellow}
\usepackage{makecell} 
\usepackage{multirow}
\usepackage{caption}
\usepackage{pifont}
\usepackage{amsmath}
\usepackage{amssymb}
\usepackage{times}
\usepackage{helvet}
\usepackage{courier}
\usepackage{newfloat}
\usepackage{gensymb}
\usepackage{listings}
\usepackage[utf8]{inputenc}
\usepackage{float} % for [H]
\usepackage[ruled,vlined]{algorithm2e}
\SetKwInput{KwIn}{Input}
\SetKwInput{KwOut}{Output}
\usepackage{colortbl}       % For row coloring
\usepackage[table,xcdraw]{xcolor} % For extended colors (gray!10, etc.)
\usepackage{array}   
\newcommand{\cmark}{\textcolor{green!60!black}{\ding{51}}}
\newcommand{\xmark}{\textcolor{red!70!black}{\ding{55}}}

\newcommand{\midsepdefault}{\aboverulesep = 0.605mm \belowrulesep = 0.984mm}

\usepackage[capitalize]{cleveref}  % load AFTER hyperref
\hypersetup{
    colorlinks=true,
    linkcolor=blue,
    filecolor=magenta,      
    urlcolor=cyan
}

\begin{document}

\title{MIC-BEV: Multi-Infrastructure Camera Bird's-Eye-View Transformer with Relation-Aware Fusion for 3D Object Detection}

\author{Yun~Zhang, Zhaoliang~Zheng, Johnson~Liu, Zhiyu~Huang$^{\ast}$, \\ Zewei~Zhou, Zonglin~Meng, Tianhui~Cai, and Jiaqi~Ma%
\thanks{All authors are with the University of California, Los Angeles (UCLA), CA 90095, USA. Email: \{yun666, zhz03, jwu7, zhiyuh, zeweizhou, meng925, tianhui, jiaqima\}@ucla.edu}%
\thanks{$^{\ast}$Corresponding author: Zhiyu Huang.}%
}



\maketitle

\begin{abstract}
Infrastructure-based perception plays a crucial role in intelligent transportation systems by providing global situational awareness and enabling cooperative autonomy. However, camera-based 3D object detection remains challenging under multi-view infrastructure setups, heterogeneous camera configurations, degraded sensing conditions, diverse road layouts, and adverse environmental conditions.
We introduce \textit{MIC-BEV}, a Transformer-based bird's-eye-view (BEV) perception framework for infrastructure-based multi-camera 3D object detection. \textit{MIC-BEV} supports a variable number of cameras with heterogeneous intrinsic and extrinsic parameters and uses a graph-enhanced fusion module to adaptively integrate multi-view image features by modeling geometric relationships between cameras and the BEV grid.
To support training and evaluation, we introduce \textit{M2I}, a synthetic dataset featuring diverse camera layouts, road geometries, environmental conditions, and sensor-degradation cases. Experiments on the simulated \textit{M2I} dataset and the real-world \textit{RoScenes} dataset show that \textit{MIC-BEV} achieves strong 3D detection performance and improved robustness under challenging conditions. Ablation studies validate the effectiveness of its key design choices, while weight analyses show that the learned fusion weights are associated with view-reliability factors and adapt to degraded camera inputs. We further report computational cost and deployment-related factors, providing additional evidence of practical feasibility and real-world deployment potential. These results indicate that \textit{MIC-BEV} is a promising framework for infrastructure-based perception. The dataset and source code are available at: \url{https://github.com/HandsomeYun/MIC-BEV}.
\end{abstract}


\section{Introduction}
\label{sec:intro}

\begin{figure}[ht]
\centering
\includegraphics[width=0.88\linewidth]{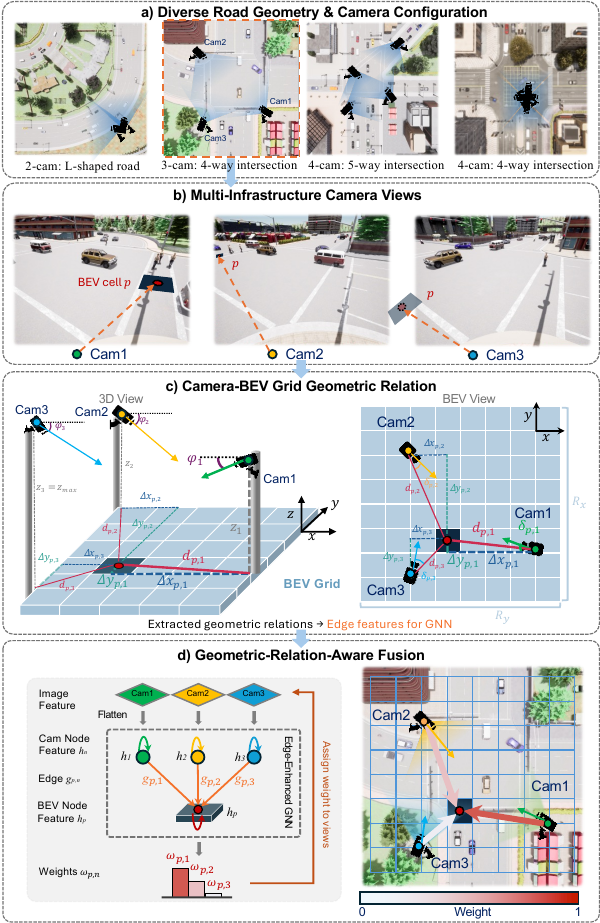}
\caption{
\textbf{Overview of the proposed MIC-BEV framework for multi-camera infrastructure perception.}
  (a) Example scenarios of infrastructure sensing with diverse road geometry and camera configuration.
  (b) Multiple cameras capture the scene from different viewpoints and project objects onto a defined BEV grid. 
  (c) MIC-BEV encodes the geometric relations between each camera and BEV cell, such as distances, angles, and height differences, to construct edge features for a graph neural network (GNN).
  (d) The GNN in MIC-BEV performs geometric relation-aware fusion, assigning importance weights to each camera for adaptive multi-view feature aggregation.
}
\label{figure1}
\vspace{-0.6cm}
\end{figure}

\IEEEPARstart{I}{nfrastructure}-based perception is a key enabler for intelligent transportation systems, providing critical support for traffic monitoring, situational awareness, and cooperative autonomy in urban environments \cite{yu2025univ2x, tang2024multi, li2022v2x}. Sensors deployed at intersections, crosswalks, and merging zones offer a significant advantage for observing traffic participants from elevated viewpoints, providing broader and more stable observations. This spatial advantage facilitates long-term monitoring and enhances the ability to detect dynamic objects \cite{bai2022infrastructure}. While LiDAR or radar sensors and multi-modal fusion with cameras provide more accurate geometric and distance information for infrastructure sensing, they often require additional hardware, calibration, synchronization, and maintenance \cite{lotscher2023assessing, jiang2023optimizing, kim2023placement}. In contrast, cameras are significantly more affordable, scalable, and easier to deploy. They also provide rich semantic information that enhances scene understanding, making them an attractive solution for large-scale infrastructure-based perception \cite{kloeker2023economic}.

While single-camera infrastructure perception systems have been extensively studied \cite{wang2024bevspread, yang2024monogae, li2022unsupervised}, their spatial coverage and robustness are inherently limited, particularly under occlusion or in complex scenes. Multi-camera infrastructure sensing addresses these limitations by integrating information from multiple viewpoints to achieve more comprehensive and resilient scene understanding \cite{xu2022cobevt}. Nonetheless, it introduces several critical challenges.
\textbf{1) Spatially distributed sensors.} Cameras deployed across large spatial distances often have overlapping fields of view with significant perspective differences and occlusions. These multi-view conditions make spatial alignment and feature fusion across views challenging. 
\textbf{2) Variability in camera configurations.} Unlike vehicle-mounted sensors that follow consistent mounting patterns, infrastructure cameras are deployed with diverse quantities, spatial layouts, orientations, fields of view (FoV), and degrees of overlap. Each intersection has a distinct design, requiring models to adapt to a wide range of installation configurations.
\textbf{3) Sensor reliability and robustness.} Infrastructure cameras may degrade, fail, or drift out of calibration over time. Consequently, perception models must remain robust to missing, corrupted, and misaligned visual inputs during real-world deployment.

To address these challenges, we propose \textbf{MIC-BEV} (\underline{\textbf{M}}ulti-\underline{\textbf{I}}nfrastructure \underline{\textbf{C}}amera Bird’s-Eye-View Transformer), an effective 3D object detection model for infrastructure-based multi-camera perception using a Bird’s-Eye-View (BEV) representation. MIC-BEV introduces a \textbf{relation-enhanced spatial cross-attention} mechanism that fuses multi-view features by jointly considering camera-specific embeddings and their geometric relations with each BEV cell through a graph neural network (GNN)~\cite{gnnsurvey}, as illustrated in \cref{figure1}. This enables adaptive weighting of information from heterogeneous camera viewpoints and improves multi-view feature aggregation across diverse camera configurations and infrastructure layouts. MIC-BEV further incorporates both map-level and object-level BEV segmentation to enhance spatial understanding. Camera masking and degradation-augmented training are employed to improve robustness to sensor degradation, camera failures, and calibration perturbations.

To overcome the scarcity of real-world datasets that capture diverse infrastructure configurations, weather conditions, and camera layouts, we introduce \textbf{M2I}, a large-scale synthetic dataset for \underline{\textbf{M}}ulti-camera, \underline{\textbf{M}}ulti-layout \underline{\textbf{I}}nfrastructure perception. M2I covers a wide range of simulated infrastructure settings with variations in camera quantity, spatial layout, heading angle, FoV, and view overlap, as well as challenging conditions such as adverse weather and varying lighting. This dataset provides a comprehensive benchmark for evaluating model generalization, robustness, and deployment feasibility. The main contributions of this paper are summarized as:
\begin{enumerate}
    \item We propose \textbf{MIC-BEV}, a BEV-based 3D object detection model for multi-camera infrastructure perception, featuring a relation-enhanced spatial cross-attention module with GNN-enhanced fusion to model camera-BEV geometric relations and perform reliability-aware multi-view aggregation.
    
    \item We introduce \textbf{M2I}, a large-scale synthetic dataset with diverse and realistic multi-camera configurations, road layouts, environmental conditions, and sensor-degradation cases, enabling systematic evaluation of generalization and robustness.
    
    \item We conduct extensive experiments on both the synthetic M2I and real-world RoScenes datasets, demonstrating state-of-the-art detection performance, robustness, and adaptive multi-view fusion across diverse simulated conditions and a real-world highway infrastructure setting.
\end{enumerate}

\section{Related Work}
\label{sec:relate}

\subsection{Camera-based BEV Perception}
BEV representations have become a dominant paradigm in camera-based 3D perception, offering a unified spatial abstraction across multi-view inputs. Early works such as OFT~\cite{roddick2018orthographicfeaturetransformmonocular} and CADDN~\cite{reading2021caddn} project monocular image features into BEV space, while Lift-Splat-Shoot~\cite{philion2020lift} lifts 2D features into 3D frustums using predicted depth distributions before accumulating them in BEV. Building on this paradigm, BEVDet~\cite{huang2021bevdet} improves efficiency for multi-view detection, BEVDepth~\cite{li2023bevdepth} enhances depth estimation through LiDAR-guided supervision, and GeoBEV~\cite{zhang2025geobev} improves geometric fidelity with radial-Cartesian sampling and centroid-aware depth supervision.

Transformer-based methods further advance BEV detection. DETR3D~\cite{wang2022detr3d} extends DETR~\cite{carion2020end} to 3D detection by sampling image features at learned 3D reference points through deformable attention, removing the need for explicit depth estimation. PETR~\cite{liu2022petr} introduces 3D positional encodings to lift 2D features into a pseudo-3D representation, while PETRv2~\cite{liu2023petrv2} adds depth supervision and improved alignment for multi-view fusion. BEVFormer~\cite{li2022bevformer} introduces learnable BEV queries with spatial and temporal deformable attention for dense BEV feature aggregation. Subsequent works further improve temporal reasoning and efficiency from different perspectives. StreamPETR~\cite{wang2023streamerpetr} performs object-centric temporal modeling by propagating historical object queries for streaming video inference. BEVDet4D~\cite{huang2022bevdet4dexploittemporalcues} extends BEVDet with temporal feature fusion, SoloFusion~\cite{park2022solo} combines short- and long-term temporal cues for multi-view stereo, and BEVNeXt~\cite{li2024bevnext} revisits dense BEV frameworks with enhanced depth modeling, temporal aggregation, and object decoding.

While these methods achieve strong results in ego-centric, vehicle-mounted scenarios, they typically assume fixed, synchronized, and well-calibrated camera layouts surrounding the ego vehicle. Infrastructure-based systems are substantially different: cameras are spatially distributed across poles, gantries, or roadside units, with diverse orientations, heights, fields of view, and overlap patterns that vary across road geometries and installation sites. This heterogeneity makes geometric alignment and multi-view fusion more challenging, motivating BEV frameworks specifically designed for infrastructure perception.

\subsection{Infrastructure-based 3D Perception and Datasets} 

Infrastructure-based perception systems often rely on LiDAR~\cite{yang2024lidar, zimmer2023real, yu2023vehicleinfrastructurecooperative3dobject} or multi-modal fusion for 3D object detection, including LiDAR-camera~\cite{zimmer2023infradet3d, bai2022transfusion, liu2022bevfusion, meng2024agentalign} and radar-camera BEV frameworks such as RCBEVDet~\cite{lin2024rcbevdet} and RCBEVDet++~\cite{lin2024rcbevdetplus}. These multi-modal methods improve robustness through sensor complementarity, where radar or LiDAR can provide geometric, range, or velocity cues under challenging visibility conditions. However, they require additional sensors, cross-modal calibration, and higher deployment complexity. Camera-only infrastructure perception remains attractive for its lower cost and rich semantics \cite{hu2022investigating,zheng2025inspe}. Early camera-based methods studied monocular 3D detection on Rope3D \cite{ye2022rope3d} and cooperative datasets \cite{yu2022dair, liu2025towards}. Methods including BEVHeight \cite{yang2023bevheight}, BEVHeight++ \cite{yang2025bevheight++}, and CoBEV \cite{shi2024cobev} enhance spatial understanding with depth-height cues, while MonoUNI \cite{jinrang2023monouni} uses normalized depth features to generalize across infrastructure and vehicle perspectives. Multi-camera infrastructure perception further expands spatial coverage and robustness. Existing datasets such as V2X-Real~\cite{xiang2024v2x} and RCooper~\cite{hao2024rcooper} explore single-intersection and corridor environments, while RoScenes~\cite{zhu2024roscenes} targets highway perception. Recent methods further address practical deployment challenges, including calibration-free fusion~\cite{fan2023calibration}, occupancy-based traffic monitoring with MC-BEVRO~\cite{vaghela2025mc}, and multi-view BEV fusion with feature-guided queries or rotation-aware embeddings in RoBEV \cite{zhu2024roscenes} and RopeBEV~\cite{jia2024ropebev}.  

However, existing multi-camera fusion strategies often aggregate views implicitly, limiting adaptation to spatially varying view reliability across camera layouts. Existing infrastructure datasets also remain limited in spatial and environmental diversity for evaluating heterogeneous deployment settings.
To address these limitations, we introduce \textbf{M2I}, a large-scale synthetic benchmark for controlled evaluation across diverse intersections, camera configurations, and environmental conditions. We propose \textbf{MIC-BEV}, a relation-aware BEV framework that models camera-BEV geometric dependencies and dynamically assigns region-specific fusion weights for adaptive multi-view aggregation.

\begin{figure*}[t]
  \centering
  \includegraphics[width=\linewidth]{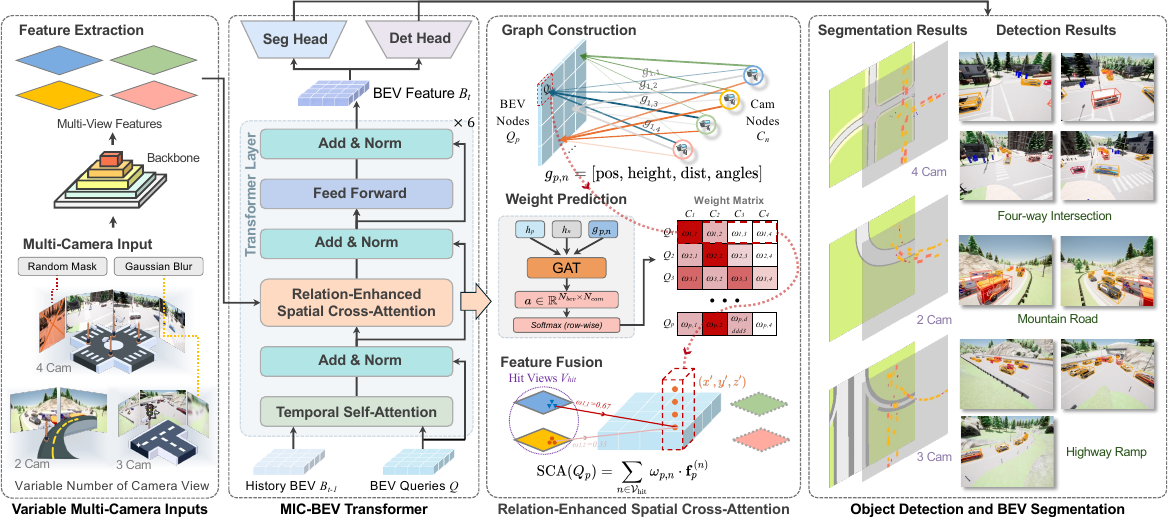}
  \caption{
    \textbf{Overview of the MIC-BEV architecture.} The model takes multi-view images from a variable number of infrastructure-mounted cameras as input and extracts features through a shared backbone. A camera masking module applies random dropout or Gaussian blur to simulate degraded views. The extracted features are fused into a BEV representation via Transformer layers with temporal self-attention and our proposed \textbf{Relation-Enhanced Spatial Cross-Attention}. GAT networks are used to dynamically assign view-dependent weights based on camera node features and geometric relations between the camera and its visible BEV cells. The resulting BEV features are used for both object detection and BEV semantic segmentation tasks.
  }
  \label{figure2}
  \vspace{-0.3cm}
\end{figure*}

\section{Methodology}
\label{sec:method}

\subsection{Problem Definition}
The objective of this work is infrastructure-mounted multi-camera 3D object detection. Given synchronized RGB images from $N_\text{cam}$ infrastructure cameras and their calibration parameters, the detector $\operatorname{Det}_{\phi}(\cdot)$ predicts 3D object instances and BEV semantic representations:
\begin{equation}
\hat{B}, \hat{M} = \operatorname{Det}_{\phi}\left( \{(I_n, E_n, K_n)\}_{n=1}^{N_\text{cam}} \right),
\end{equation}
where $I_n$ denotes the image captured by the $n$-th camera, and $E_n$ and $K_n$ denote the corresponding extrinsic and intrinsic matrices. $N_\text{cam}$ may vary across scenes, reflecting heterogeneous infrastructure deployments with different numbers of cameras, poses, and fields of view.

The detection branch outputs a set of 3D bounding boxes $\hat{B}$ in a unified coordinate system, where each box is parameterized as $\hat{B_i} =(x,y,z,l,w,h,\psi)$, representing the object center, box dimensions, and yaw orientation.
In parallel, the segmentation branch predicts BEV semantic maps
$\hat{M} = [\hat{M}_{\text{map}}, \hat{M}_{\text{object}}]$, where $\hat{M}_{\text{map}}$ represents static scene semantics (e.g., road, crosswalk) and $\hat{M}_{\text{object}}$ captures dynamic object occupancy (e.g., vehicle, pedestrian). Joint optimization of detection and segmentation enables the model to reason about both scene layout and object-level spatial structure.

\textbf{MIC-BEV} addresses this problem through a unified BEV Transformer framework for heterogeneous infrastructure camera settings. As shown in \cref{figure2}, MIC-BEV extracts multi-view image features from a variable number of cameras and aggregates them into BEV space using temporal modeling and relation-enhanced spatial cross-attention. Its key component is a GNN-enhanced fusion module, which models camera-BEV geometric relations and assigns reliability-aware view weights for each BEV cell. This enables MIC-BEV to adaptively emphasize informative camera views while suppressing degraded, occluded, or geometrically unfavorable observations. The fused BEV representation is then used for joint 3D object detection and BEV semantic segmentation.

%%%%%%%%%%%%%%%%%%%%%%%%%%%%%%%%%%%%%%%%%%%%%%%%%%%%%%%%
\subsection{Variable Multi-Camera Inputs}

MIC-BEV supports a variable number of input cameras. When fewer than the maximum supported cameras are available, missing views are padded with zero-valued images and identity calibration matrices, which are excluded from subsequent attention and graph-based fusion operations.
To improve robustness under imperfect sensing conditions, we introduce view-masking and calibration-noise augmentations during training to simulate sensor degradation, camera failure, and calibration errors. With probability $p_m$, a randomly selected camera view is either masked or Gaussian blurred. In addition, with probability $p_c$, calibration perturbations are applied to one camera by injecting random noise into its extrinsic parameters. These augmentations encourage the model to learn more adaptive multi-view fusion behavior under incomplete, degraded, or miscalibrated sensing conditions.

\subsection{MIC-BEV Transformer}
\label{sec:relation_enhanced_attention}

The process of encoding multi-camera features into BEV space in MIC-BEV is summarized in \cref{alg:micbev_forward}. The Transformer model is designed to handle heterogeneous camera configurations while preserving spatial and temporal consistency.

\subsubsection{\textbf{Encoder and BEV Queries}}
We employ a ResNet-101~\cite{he2016deep} backbone coupled with a Feature Pyramid Network (FPN)~\cite{lin2017feature} to extract multi-scale features from each camera image. The BEV representation is defined as a 2D grid anchored to the ground plane and centered at the scene. We initialize a learnable tensor $\mathbf{Q} \in \mathbb{R}^{H_{\text{bev}} \times W_{\text{bev}} \times C}$ to represent the grid, where $H_{\text{bev}}$ and $W_{\text{bev}}$ denote the grid resolution, and $C$ is the feature dimension. Each cell $Q_p \in \mathbb{R}^{C}$ serves as a latent query corresponding to a grid location $p=(x,y)$ in the BEV space.
These BEV queries interact with multi-view image features via spatial cross-attention and are iteratively refined to capture spatial cues encoded by the mounted cameras.

\begin{algorithm}[ht]  
\caption{Forward Pass of MIC-BEV Encoding}  
\label{alg:micbev_forward}  
\KwIn{$\{\mathbf{B}_{t-k}\}_{k=1}^{T}$: past $T$ frames BEV history queue;\\  
\quad\quad $\{\mathbf{F}^{(n)}\}_{n=1}^{N_\text{cam}}$: multi-view image features;\\  
\quad\quad $\{\mathbf{K}_n,\mathbf{E}_n\}_{n=1}^{N_\text{cam}}$: intrinsics/extrinsics;\\  
\quad\quad $L$: number of Transformer layers;\\  
\quad\quad $N_{\text{ref}}$: number of anchor heights}  
\KwOut{$\mathbf{B}_{\text{out}}$: current-frame BEV features}  
\BlankLine  
  
\textbf{1. Generate reference points}

$\mathbf{r}^{3\!D} \gets \textsc{GetRefPoints3D}(H_{\text{bev}}, W_{\text{bev}}, N_{\text{ref}})$ 
\tcp*{Shape: $[N_{\text{ref}}, H_{\text{bev}}, W_{\text{bev}}, 3]$}  

\textbf{2. Project 3D anchors to camera views (once)}

$\{(\mathbf{u}_n,\mathbf{m}_n)\}_{n=1}^{N_{\text{cam}}} \gets \textsc{PointSampling}(\mathbf{r}^{3\!D}, \{\mathbf{K}_n,\mathbf{E}_n\})$ 
\tcp*{$\mathbf{m}_n$ masks invalid points}  
  
\textbf{3. Initialize temporal references (static cameras)}

\eIf{$T>0$}{  
    $\mathbf{B}_{\text{prev}} \gets \textsc{Concat}(\{\mathbf{B}_{t-k}\}_{k=1}^{T})$
}{  
    $\mathbf{B}_{\text{prev}} \gets \textsc{ZerosLike}(\mathbf{Q})$
}

\textbf{4. Transformer layers}

$\mathbf{B}_0 \gets \mathbf{Q}$ \tcp*{Initialize with learnable BEV queries}
\For{$\ell = 1$ \KwTo $L$}{
    \tcp{Temporal Self-Attention (TSA)}
    $\mathbf{B}_\ell \gets \textsc{TSA}(\mathbf{B}_{\ell-1}, \mathbf{B}_{\text{prev}})$\;
    $\mathbf{B}_\ell \gets \textsc{Norm}(\mathbf{B}_\ell)$\;

    \tcp{Relation-Enhanced Spatial Cross-Attention (ReSCA)}
    $\mathbf{B}_\ell \gets \textsc{ReSCA}(\mathbf{B}_\ell, \{\mathbf{F}^{(n)}\}, \mathbf{r}^{3\!D}, \{\mathbf{u}_n\}, \{\mathbf{m}_n\})$ %\tcp*{Per-view weights via bipartite GAT (\cref{alg:cam-bev-gat})}
    $\mathbf{B}_\ell \gets \textsc{Norm}(\mathbf{B}_\ell)$\;

    \tcp{Feed-Forward Network (FFN)}
    $\mathbf{B}_\ell \gets \textsc{FFN}(\mathbf{B}_\ell)$\;
    $\mathbf{B}_\ell \gets \textsc{Norm}(\mathbf{B}_\ell)$\;
}
\Return $\mathbf{B}_{\text{t}}=\mathbf{B}_L$  
\end{algorithm}
%%%%%%%%%%%%%%%%%%%%%%%%%%%%%%%%%%%%%%%%%%%%%%%%%%%%%%%%

\subsubsection{\textbf{Temporal Self-Attention}} 
To capture object motion dynamics, temporal self-attention (TSA) allows current BEV queries $Q$ to attend to the previous BEV map $B_{t-1}$. In static infrastructure setups, this is simplified without ego-motion compensation. Incorporating temporal context helps recover occluded or intermittently visible objects and stabilizes predictions over consecutive frames.

\subsubsection{\textbf{Relation-Enhanced Spatial Cross-Attention}}
Given a set of multi-view camera feature maps $\{F^{(n)}\}_{n=1}^{N_\text{cam}}$, the proposed Relation-Enhanced Spatial Cross-Attention (ReSCA) aggregates them into a unified BEV representation $F\in\mathbb{R}^{C \times H_{\text{bev}} \times W_{\text{bev}}}$. For each BEV query $Q_p$ located at $p$ in the BEV grid, we generate a vertical stack of $N_{\text{ref}}$ 3D reference points $\mathbf{r}_{p,j} = (x, y, z_j)$ using a predefined set of anchor heights $\{z_j\}_{j=1}^{N_{\text{ref}}}$. These pillars help capture semantic features across different heights.
Each 3D reference point $\mathbf{r}_{p,j}$ is projected onto the $n$-th camera view as 2D coordinates $\mathbf{u}_{p,j}^{(n)}$. Only camera views where the projected points fall within valid image bounds are included in the hit-view set $\mathcal{V}_{\text{hit}}$.

For each hit view $n \in \mathcal{V}_{\text{hit}}$, we apply deformable attention ($\mathrm{DeformAttn}$) \cite{zhu2020deformable} around the projected locations $\{\mathbf{u}_{p,j}^{(n)}\}_{j=1}^{N_{\text{ref}}}$ of 3D reference points associated with BEV query $Q_p$. This produces a per-view feature $\mathbf{f}_p^{(n)} \in \mathbb{R}^C$.

%%%%%%%%%%%%%%%%%%%%%%%%%%%%%%%%%%%%%%%%%%%%%%%%%%%%%%%%
\textbf{Graph Construction}.
To adaptively weight the contribution of each camera view for every BEV cell, we formulate multi-view fusion as a bipartite graph attention problem. The conventional uniform averaging of multi-view features ignores how \emph{informative} or \emph{reliable} each view is, especially under occlusions or perspective bias. We therefore learn the fusion weights $\omega_{p,n}$ using a geometry- and content-aware Graph Attention Network (GAT)~\cite{velivckovic2017graph}. 

We construct a bipartite graph $\mathcal{G} = (\mathcal{V}_{\text{cam}}, \mathcal{V}_{\text{bev}}, \mathcal{E})$, where each camera node $C_n \in \mathcal{V}_{\text{cam}}$ represents a pooled image feature embedding from camera $n$, and each BEV grid cell node $Q_p \in \mathcal{V}_{\text{bev}}$ is represented by a BEV query located at $p$. The node features are defined as:
\begin{equation}
\mathbf{h}_p = Q_p \in \mathbb{R}^C \quad \text{for BEV nodes}, 
\end{equation}
\begin{equation}
\mathbf{h}_n = \frac{1}{K} \sum_{k=1}^{K} f_{n,k} \in \mathbb{R}^C \quad \text{for camera nodes},
\end{equation}
where \( K = H \times W \) is the number of tokens from the camera feature map \( F^n \in \mathbb{R}^{C \times H \times W} \), with $H$ and $W$ denoting the height and width of the feature map, respectively. \( f_{n,k} \) denotes the \( k \)-th token feature from camera \( n \).

Edges $\mathcal{E}$ are directed from cameras to visible BEV nodes, \(\mathcal{E} = \{(n, p) \mid Q_p \text{ is visible from camera } C_n\}\). Each edge \((n \to  p)\) is annotated with a geometric relation descriptor \(\mathbf{g}_{p,n} \in \mathbb{R}^8\), consisting of:
\begin{equation}
\label{eq:geometry}
\begin{aligned}
\mathbf{g}_{p,n} = \Big[
&\tfrac{\Delta x_{p,n}}{R_x},\ \tfrac{\Delta y_{p,n}}{R_y},\ \tfrac{z_n}{z_{max}},\ 
\tfrac{||\mathbf{d}_{p,n}||_2}{\sqrt{R_x^2 + R_y^2}}, \\
&\cos\delta_{p,n},\ \sin\delta_{p,n},\ 
\cos\varphi_n,\ \sin\varphi_n
\Big],
\end{aligned}
\end{equation}
where \(\mathbf{d}_{p,n} =(\Delta x_{p,n}, \Delta y_{p,n}) = (x_p - x_n, y_p - y_n)\) is the 2D planar offset between the BEV cell center and the camera center. $R_x$ and $R_y$ are normalization constants corresponding to the sensing range in $x$ and $y$ directions, used to scale spatial offsets to a consistent range within $[-1, 1]$. Similarly, \(z_n\) is the camera's height, normalized by the maximum camera height \(z_{max}\). \(\delta_{p,n}\) is the heading difference between the camera's yaw and the angle from camera $n$ to the BEV cell at location $p$, and \(\varphi_n\) is the pitch angle of camera $n$.

The geometric encoding in \cref{eq:geometry} is designed to describe physical factors that influence camera-view reliability in infrastructure perception. The relative offsets and distance terms encode spatial coverage and scale variation, camera height captures viewpoint differences and depth uncertainty, and heading/pitch terms characterize the alignment between each camera and BEV region. We represent angular quantities with sine and cosine functions to avoid discontinuities near $\pm\pi$ and provide smooth relation features for attention-based weighting.

We use graph attention because camera reliability is both location-dependent and relation-dependent: the most informative camera can vary across BEV cells due to visibility, overlap, and perspective distortion. The bipartite camera-to-BEV graph explicitly models these camera-cell relations, while the attention mechanism learns adaptive fusion weights instead of relying on uniform averaging or fixed camera preferences.

\begin{algorithm}[ht]
\caption{\small Camera-to-BEV Fusion Weights via GAT}
\label{alg:cam-bev-gat}
\KwIn{
$\mathbf{Q}\!\in\!\mathbb{R}^{B\times N_q\times C}$: BEV queries; 
$\{\mathbf{F}^{(n)}\!\in\!\mathbb{R}^{B\times C\times H\times W}\}_{n=1}^{N_\text{cam}}$: camera features;\\
$\mathbf{M}\!\in\!\{0,1\}^{B\times N_q\times N_\text{cam}}$: BEV visibility mask; 
$(\mathbf{P},\boldsymbol{\psi},\boldsymbol{\varphi})$: camera positions, yaw, pitch; 
$f_\theta$: GATv2
}
\KwOut{$\mathbf{W}\!\in\!\mathbb{R}^{B\times N_q\times N_\text{cam}} $: fusion weights}
\smallskip

\textbf{1. Node features:}
Mean-pool image tokens to get camera embeddings  
\(\mathbf{C}_b[n]\!=\!\frac{1}{HW}\!\sum_{h,w}\!\mathbf{F}^{(n)}_{b,:,h,w}\);
use \(\mathbf{Q}_b\) as BEV node features.

\textbf{2. Edges:}
From the visibility mask  
\(\mathcal{E}_b=\{(n\!\to\!q)\mid \mathbf{M}_{b,q,n}=1\}\)
(connect visible cameras to BEV cells).

\textbf{3. Edge attributes:}
For each edge $(b,q,n)$ compute geometry vector  
\[
\mathbf{g}_{b,q,n}
=[\tfrac{\Delta x}{R_x},\tfrac{\Delta y}{R_y},\tfrac{z_n}{z_{\max}},
\tfrac{\|\mathbf{d}\|_2}{\sqrt{R_x^2+R_y^2}},
\cos\delta,\sin\delta,\cos\varphi,\sin\varphi].
\]

\textbf{4. GAT propagation:}
\(\mathbf{S}_b=f_\theta(\mathbf{Q}_b,\mathbf{C}_b,\mathcal{E}_b,\mathbf{g}_b)
\in\mathbb{R}^{N_q\times N_\text{cam}}\).

\textbf{5. Normalization:}
Mask invalid edges  
(\(\mathbf{S}_{b,q,n}\!=\!-\infty\) if \(\mathbf{M}_{b,q,n}\!=\!0\)),  
Then apply softmax across cameras:  
\(\mathbf{W}_{b,q,:}=\texttt{softmax}_n(\mathbf{S}_{b,q,:})\).

\Return $\mathbf{W}$.
\end{algorithm}

\textbf{Weight Prediction}.
We employ a GAT network $f_\theta$ to model the interaction between each BEV node, camera node, and their geometric relation, as detailed in \cref{alg:cam-bev-gat}. Specifically, $f_\theta$ is instantiated as a GATv2~\cite{brody2021attentive} with three layers, four attention heads per layer, and 128 hidden units.  

The network outputs a scalar logit $s_{p,n}$ for each relation pair $(n \to p)$, representing the importance weight of the camera node $n$ for the BEV cell $p$:
\begin{equation}
\label{eq:gat-score}
s_{p,n} = f_\theta(\mathbf{h}_p,\, \mathbf{h}_n,\, \mathbf{g}_{p,n}).
\end{equation}

The logits are reshaped to $(B, N_q, N_\text{cam})$ and normalized across cameras using a visibility-aware softmax. For non-visible cameras, we set $s_{p,n}\!\leftarrow\!-\infty$ to exclude them. The final fusion weights $\omega_{p,n}$ are then computed as:
\begin{equation}
\omega_{p,n} = 
\frac{\exp(s_{p,n})}
{\sum_{n' \in \mathcal{V}_{\text{hit}}} \exp(s_{p,n'})}.
\end{equation}

This formulation enables adaptive weighting across multiple camera views, where each BEV cell dynamically attends to the most informative and geometrically relevant viewpoints.

\textbf{Feature Fusion}.
The final BEV feature is computed by fusing all visible views with learned weights $\omega_{p, n}$:
\begin{equation}
\label{fusion}
\begin{aligned}
\mathrm{ReSCA}(Q_p) = \sum_{n \in \mathcal{V}_{\text{hit}}} \omega_{p , n} \cdot \mathbf{f}_p^{(n)}, \,
\sum_{n} \omega_{p,n} = 1, &\\
\mathbf{f}_p^{(n)} = \sum_{j=1}^{N_{\text{ref}}} \mathrm{DeformAttn}(Q_p,\, \mathbf{u}_{p,j}^{(n)},\, F^{(n)}). &
\end{aligned}
\end{equation}

This weighted aggregation allows each BEV cell to integrate complementary information from multiple camera views according to the learned fusion weights. By emphasizing informative observations while reducing the influence of unreliable views, ReSCA produces a more robust and consistent BEV representation across diverse camera configurations.

\subsection{Object Detection and BEV Segmentation}

The BEV Transformer outputs a shared BEV feature map 
$B_t \in \mathbb{R}^{C \times H_{\text{bev}} \times W_{\text{bev}}}$, which supports two parallel tasks: 3D object detection and BEV segmentation. 

For object detection, MIC-BEV adopts a DETR-style decoder~\cite{carion2020end} with $N_{\text{obj}}$ object queries. Each query predicts an object category and its corresponding 3D bounding box. The detection branch is trained with bipartite matching, using focal loss for classification and L1 loss for box regression. 

For BEV segmentation, lightweight convolutional decoders are attached to the shared BEV feature map to predict static map regions and dynamic object masks. These branches are supervised by rasterized HD maps and object annotations using cross-entropy loss. The overall training objective combines detection and segmentation losses, with the segmentation loss weighted by $\lambda=2$.

\begin{table*}[ht]
  \centering
  \caption{
    Comparison of infrastructure perception datasets.
    Weather and map availability are key attributes, while variation captures diversity across layouts, locations, and fields of view. Prior datasets often have fixed camera setups and narrow scene types, whereas M2I offers broader diversity and richer supervision.
  }
  \resizebox{\linewidth}{!}{
  \begin{tabular}{@{}l c c c c c c c c c c@{}}
    \toprule
    \textbf{Dataset} & \textbf{Type} & \textbf{Year} & \textbf{Frames} & \textbf{Boxes} & \textbf{\# Cams} & \textbf{Extreme Weather} & \textbf{Map} & \textbf{Vary Layouts} & \textbf{Vary Locations} & \textbf{Vary FoVs} \\
    \midrule
    V2X-Sim-I~\cite{li2022v2x}          & Sim  & 2022 & 60K   & 26.6K  & 4       &  \xmark   &  \xmark & \xmark  &  \cmark  & \xmark  \\
    V2XSet-I~\cite{xu2022v2x}           & Sim  & 2022 & 44K   & 233K   & 4       &  \xmark  &  \xmark &  \xmark  &   \cmark    & \xmark  \\
    DAIR-V2X-I~\cite{yu2022dair}        & Real & 2022 & 10K   & 493K   & 1       &  \xmark  & \xmark  & \xmark  &  \cmark  & \xmark   \\
    Rope3D~\cite{ye2022rope3d}          & Real & 2022 & 50K   & 1.5M   & 1       & \xmark   &\xmark & \xmark  & \cmark  & \xmark  \\
    V2X-Real-I~\cite{xiang2024v2x}      & Real & 2023 & 171K  & 1.2M   & 4       & \xmark  & \xmark  & \xmark  & \xmark  & \xmark \\
    TUMTraf-I~\cite{waltertumtraf} & Real & 2023 & 2.4K & 57.4K & 2 & \xmark & \cmark & \xmark & \xmark  & \cmark \\
    V2X-Seq-I~\cite{xiang2024v2x}       & Real & 2023 & 39K   & 464K   & 2       & \xmark  & \cmark & \xmark  & \xmark  & \xmark  \\
    V2X-PnP-I~\cite{zhou2024v2xpnp}     & Real & 2024 & 208K  & 1.45M  & 4       & \xmark  & \cmark & \xmark  & \xmark  & \xmark \\
    TUMTraf-V2X ~\cite{zimmer2024tumtraf}     & Real & 2024 & 2.0K & 30K  & 4       & \xmark  & \cmark  & \xmark  & \xmark & \xmark  \\
    RCooper~\cite{hao2024rcooper}       & Real & 2024 & 50K   & 242K   & 2-4    & \xmark  & \xmark & \cmark  & \xmark  & \xmark \\
    RoScenes~\cite{zhu2024roscenes}     & Real & 2024 & 215K  & \textbf{21.1M}  & 8       & \xmark  & \xmark & \cmark  & \cmark  & \cmark  \\
    \rowcolor{blue!7}
    \textbf{M2I (Ours)}                 & Sim  & 2025 & \textbf{278K} & {11.6M} & 1-4 & \textbf{\cmark} & \textbf{\cmark} & \textbf{\cmark} & \textbf{\cmark} & \textbf{\cmark} \\
    \bottomrule
  \end{tabular}
  \midsepdefault
  }
  \label{tab:v2x_datasets}
\end{table*}

\begin{figure*}
    \centering
    \includegraphics[width=0.99\linewidth]{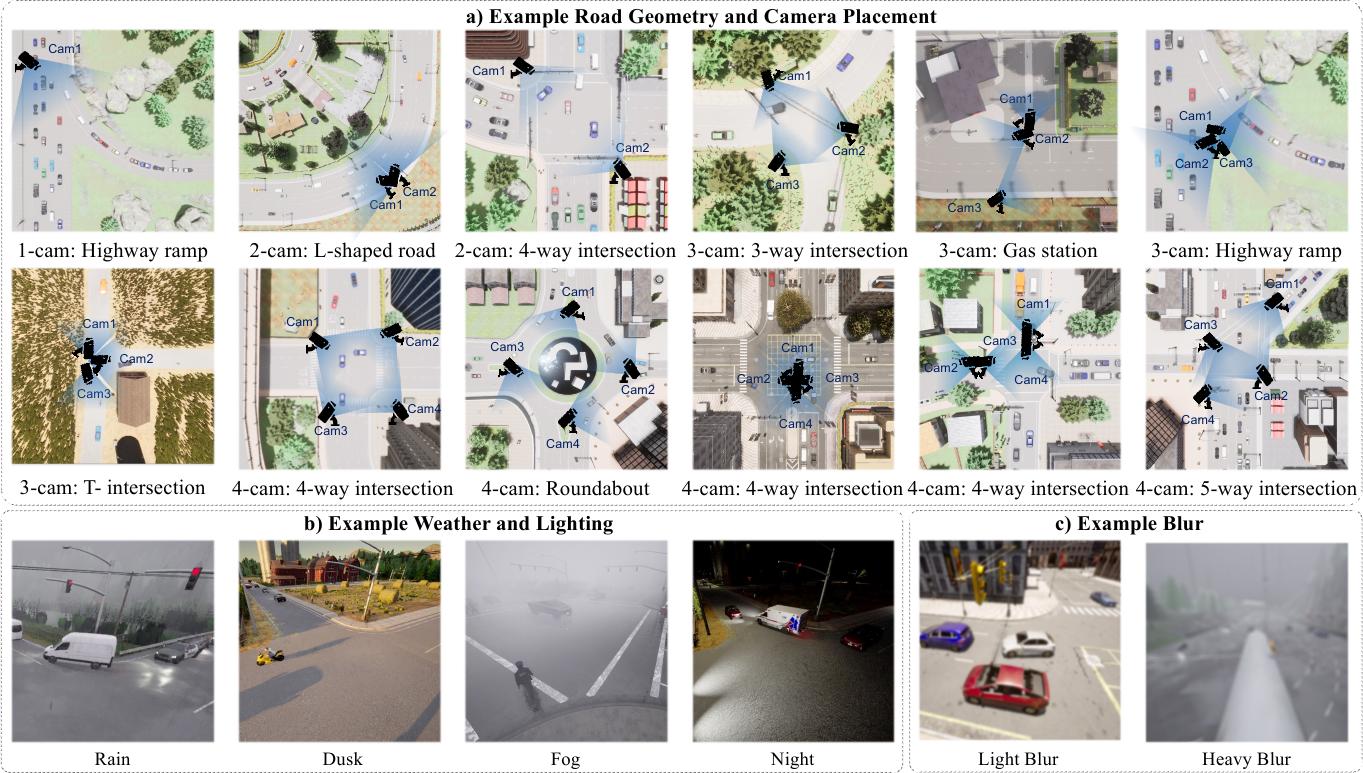}
    \caption{
        \textbf{Scene diversity in the M2I dataset.}
        (a) Examples of camera placement configurations across various intersection types, ranging from single-camera highway ramps to multi-camera urban intersections.
        (b) Diverse weather and illumination conditions, including rain, fog, dusk, and night scenes.
        (c) Different levels of image blur representing sensor degradation from light to heavy.
        These variations illustrate the broad coverage and realism of M2I for infrastructure perception.
        }
    \label{fig:examplem2i}
\end{figure*}

\section{Dataset}

\subsection{Limitations of Existing Infrastructure Sensing Datasets}

Existing datasets only partially reflect the complexity of real-world infrastructure perception, hindering the generalization of learned models across diverse deployment scenarios. 

\textbf{Limited camera configuration.} 
Existing infrastructure datasets exhibit limited variability in camera setups, often relying on single static viewpoints or fixed multi-camera arrangements. Monocular camera datasets such as Rope3D~\cite{ye2022rope3d} and DAIR-V2X-I~\cite{yu2022dair} use a single stationary camera, leading to narrow spatial coverage and no possibility for cross-view fusion. Multi-camera datasets, including RCooper~\cite{hao2024rcooper}, V2X-Real~\cite{xiang2024v2xreal}, and V2X-PnP~\cite{zhou2024v2xpnp}, place cameras at a single intersection with fixed geometry, providing no variation in setup across different environments. In both cases, camera layouts, heights, and orientations remain homogeneous, limiting the model's adaptability to diverse road geometries and real-world deployment conditions.

\textbf{Limited scene diversity.} 
Beyond camera setup, existing datasets capture only a narrow range of spatial and contextual environments. Intersection-focused datasets such as RCooper~\cite{hao2024rcooper}, V2X-Real~\cite{xiang2024v2xreal}, and V2X-PnP~\cite{zhou2024v2xpnp} are recorded at a single, fixed intersection, where the same geometry, layout, and surroundings appear throughout the dataset. Similarly, highway-focused datasets like RoScenes~\cite{zhu2024roscenes} are limited to one highway with linear road topology and low interaction density, lacking complex cross-traffic or vulnerable road users such as pedestrians and cyclists. As a result, these benchmarks offer minimal variation in spatial layout or contextual diversity, constraining the development of perception models capable of generalizing across diverse real-world road environments.

\textbf{Limited dynamic conditions.}
Most existing infrastructure datasets are collected primarily under clear daytime conditions, providing limited variation in illumination or weather. Adverse environments such as heavy rain and dense fog are typically excluded, while dusk and night scenes appear only sparsely. Consequently, models trained on these datasets often struggle to handle realistic illumination changes, sensor noise, and dynamic temporal variations, which affect perception robustness.

\begin{figure*}[ht]
    \centering
    \includegraphics[width=0.99\linewidth]{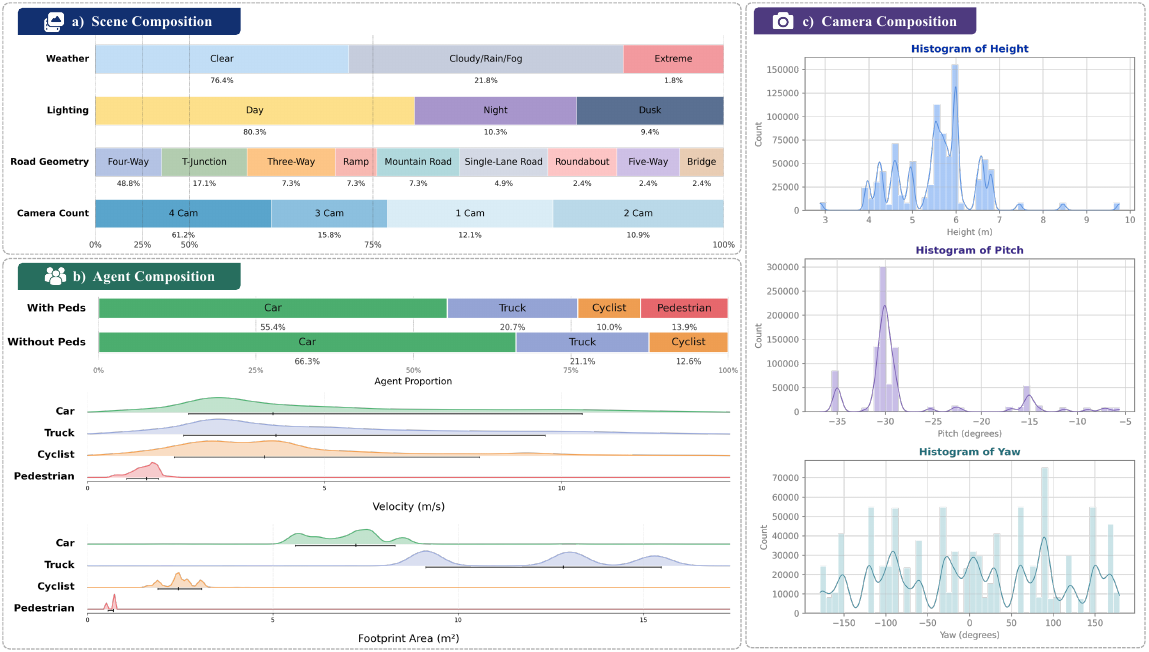}
     \caption{\textbf{M2I Dataset Composition.} (a) Scene composition showing splits by weather, lighting, road geometry, and camera count. (b) Agent composition illustrating class proportions, velocity profiles, and footprint areas across object types. (c) Camera composition showing histograms of camera height, pitch, and yaw across all scenes. These statistics demonstrate the diversity and balance of the M2I dataset for multi-camera perception.}
    \label{fig:m2isplit}
    \vspace{-0.3cm}
\end{figure*}

\subsection{M2I Dataset}

To address the limited diversity of existing infrastructure datasets, we introduce \textbf{M2I} (\textbf{M}ulti-camera \textbf{M}ulti-layout \textbf{I}nfrastructure), a large-scale benchmark for 3D perception in heterogeneous roadside environments. As summarized in \cref{tab:v2x_datasets}, M2I contains 278K frames, 927K images, and 11.6M annotated bounding boxes with synchronized RGB cameras, LiDAR, 3D annotations, and semantic BEV maps. It contains 1,103 scenario clips, each with 200-300 frames at 10 Hz and an average of 40 road users. The dataset supports multiple perception tasks, including 3D object detection, semantic BEV segmentation, multi-object tracking, and trajectory prediction. We split the clips into training, validation, and test sets with a ratio of 7:1:2.

\textbf{Scene Diversity.} 
Collecting large-scale infrastructure data across varied real-world environments is financially and logistically challenging. To enable controlled yet diverse evaluation, M2I is built upon CARLA~\cite{dosovitskiy2017carla} and spans 41 locations from 7 towns, covering a broad range of roadside environments from residential neighborhoods to rural mountainous regions. As illustrated in \cref{fig:examplem2i}(a), the dataset includes 9 categories of road geometries, including intersections, T-junctions, roundabouts, highway ramps, rural roads, mountain roads, and bridges. Within each category, scenes differ in intersection scale, lane topology, traffic signal placement, and pedestrian crossing design. Each location is also paired with a semantic BEV map, providing structural context for downstream perception tasks.

\textbf{Camera Configuration Diversity.}
M2I captures realistic variability in infrastructure-mounted camera deployments inspired by real-world traffic surveillance systems. Each scene contains 1-4 cameras selected from 12 configurations, spanning mounting locations on poles, traffic lights, and building facades with diverse heights, orientations, and fields of view. As shown in \cref{fig:m2isplit}(c), camera heights range from 3~m to over 10~m, pitch angles vary from $-5^\circ$ to $-35^\circ$, and yaw angles cover the full $360^\circ$. The layouts further include realistic intra-category variations. For example, cameras at four-way intersections may be distributed across all corners~\cite{zimmer2024tumtraf}, aligned along opposite corners~\cite{xiang2024v2xreal}, or concentrated near the intersection center~\cite{hao2024rcooper}. These configurations induce heterogeneous viewpoints, asymmetric visibility, and different occlusion patterns, making M2I suitable for evaluating multi-camera fusion under non-uniform sensor layouts.

\textbf{Environmental Diversity.} 
M2I covers diverse weather and illumination conditions, including rain, fog, dawn, dusk, and nighttime scenes, as illustrated in \cref{fig:examplem2i}(b). These conditions introduce substantial perception challenges such as reduced visibility, glare, motion blur, and contrast degradation. To further approximate practical sensing conditions, M2I also simulates image blur, temporary sensor dropout, and partial occlusion, as shown in \cref{fig:examplem2i}(c). Traffic patterns vary across environments, ranging from sparse highway traffic dominated by fast-moving vehicles to dense urban environments with frequent multi-agent interactions. Each environment includes low-, medium-, and high-density traffic settings, enabling evaluation under diverse congestion and occlusion conditions.

\textbf{Object Composition Diversity.}
As shown in \cref{fig:m2isplit}(b), M2I covers four major road-user categories: cars, trucks, cyclists, and pedestrians. Each category further contains diverse subtypes and appearances, such as different vehicle models, two-wheeler types, and pedestrian attributes, including adults and children, reflecting the heterogeneity of real-world traffic participants. The object composition is designed to resemble large-scale autonomous driving datasets such as nuScenes~\cite{caesar2020nuscenes} and Waymo~\cite{sun2020scalability}. Object scale, spatial distribution, and motion patterns vary across categories and environments, supporting a comprehensive evaluation of 3D perception under complex roadside traffic scenarios.

\subsection{RoScenes Dataset}

We further evaluate on the RoScenes dataset~\cite{zhu2024roscenes}, the largest real-world multi-view roadside perception benchmark to date. RoScenes complements M2I by enabling validation under real-world sensing conditions, including sensor noise, calibration drift, and long-tail data imbalance that are difficult to simulate accurately. The dataset comprises multiple 800-meter highway segments monitored by eight synchronized infrastructure-mounted cameras. It contains over 21 million 3D bounding box annotations collected under varying traffic densities. Each scene is annotated with four vehicle categories (car, van, bus, and truck) and is recorded under both daytime and nighttime conditions in clear weather.

Although RoScenes exhibits more limited environmental diversity than M2I, it captures realistic sensing artifacts arising from real-world infrastructure deployment, including motion blur, exposure variation, long-range appearance degradation, and calibration imperfections. Consequently, RoScenes primarily validates performance under real-world highway infrastructure settings, whereas complex urban intersections, vulnerable road users, and diverse adverse weather conditions are evaluated using the M2I benchmark. For evaluation, the dataset additionally defines hard and easy subsets based on scene-level occlusion statistics, corresponding to the top 10\% most occluded and bottom 10\% least occluded scenes, respectively. This protocol enables controlled evaluation under varying levels of visual complexity and occlusion.

\section{Experiments}
\subsection{Experimental Setup}

\textbf{Implementation Details}.
All models use a ResNet-101 backbone with deformable convolutions (ResNet101-DCN) as the image encoder. Training is performed for 10 epochs using the AdamW optimizer with an initial learning rate of $2\times10^{-4}$ and a cosine annealing schedule on 8 NVIDIA L40S GPUs with mixed-precision training. For M2I, input images are resized to $800\times600$ with a $200\times200$ BEV grid covering $[-51.2, 51.2]$\,m along both axes. For RoScenes, images remain at $1920\times1080$ resolution with a $1000\times250$ BEV grid spanning $[-400,400]$\,m in X and $[-100,100]$\,m in Y. The BEV origin is aligned with either the intersection center or the reference camera position, depending on the scene layout. Since RoScenes does not provide semantic maps or LiDAR supervision, BEV segmentation is disabled, and only 3D object detection is evaluated. To assess robustness under degraded sensing conditions, we additionally evaluate M2I using random view corruption, where one valid camera view is randomly masked or Gaussian-blurred during testing. Additional hyperparameter settings are available in the open-source code repository.

\textbf{Evaluation Metrics}.
We adopt mean Average Precision (mAP) and a modified nuScenes Detection Score (NDS) \cite{caesar2020nuscenes} based on the available evaluation metrics. Following the nuScenes protocol, mAP is computed as:
\begin{equation}
\text{mAP} = \frac{1}{|\mathcal{C}| \cdot |\mathcal{D}|} \sum_{c \in \mathcal{C}} \sum_{d \in \mathcal{D}} \text{AP}_{c,d},
\end{equation}
where $\mathcal{C}$ is the set of object classes, $\mathcal{D}$ is the set of center-distance thresholds, and $\text{AP}_{c,d}$ is the average precision for class $c$ at distance threshold $d$.

Since velocity and attribute metrics are not evaluated, NDS combines mAP with three true positive metrics: mean Average Translation Error (mATE), mean Average Scale Error (mASE), and mean Average Orientation Error (mAOE):
\begin{equation}
\text{NDS} = \frac{1}{6} \left[3 \cdot \text{mAP} + \sum_{\text{mTP}} (1 - \min(1, \text{mTP})) \right],
\end{equation}
where $\text{mTP} \in \{\text{mATE}, \text{mASE}, \text{mAOE}\}$.

\textbf{Baselines}.
We compare our method against representative BEV-based 3D detection frameworks, including Lift-Splat-Shoot (LSS) \cite{philion2020lift}, BEVFormer \cite{li2022bevformer}, DETR3D \cite{wang2022detr3d}, PETR \cite{liu2022petr}, PETRv2 \cite{liu2023petrv2}, GeoBEV \cite{zhang2025geobev}, BEVNeXt \cite{li2024bevnext}, StreamPETR \cite{wang2023streamerpetr}, and UVTR \cite{li2022uvtr}. These methods cover diverse feature lifting and BEV aggregation strategies while operating under the same camera-only input setting, providing a strong benchmark for infrastructure-based 3D perception. To support heterogeneous camera configurations, all baselines are adapted using the same variable-view padding protocol as MIC-BEV. RoBEV and RopeBEV are excluded from M2I due to unavailable public implementations. The RoScenes evaluation protocol follows RopeBEV~\cite{jia2024ropebev}.

%%%%%%%%%%%%%%%%%%%%%%%%%%%%%%%%%%%%%%%%%%%%%%%%%%%%%%%%%%%%%%%%%%%%%%%%%%%%%%%%%%%%%%

\begin{table*}[t]
  \centering
    \caption{Comparison of 3D object detection performance on the {M2I} test set under three evaluation conditions: {Normal}, {Robust} (e.g., sensor dropout or degradation), and {Extreme Weather} (e.g., heavy rain or fog).}
  \begin{tabular}{@{}l|cccc|cccc|cccc@{}}
    \toprule
    \multirow{2}{*}{Method} &
    \multicolumn{4}{c|}{\textbf{Normal}} &
    \multicolumn{4}{c|}{\textbf{Robust}} &
    \multicolumn{4}{c}{\textbf{Extreme Weather}} \\
    \cmidrule(lr){2-5} \cmidrule(lr){6-9} \cmidrule(lr){10-13}
    & \textbf{mAP} $\uparrow$ & \textbf{NDS} $\uparrow$ & mATE $\downarrow$ & mASE $\downarrow$
    & \textbf{mAP} $\uparrow$ & \textbf{NDS} $\uparrow$ & mATE $\downarrow$ & mASE $\downarrow$
    & \textbf{mAP} $\uparrow$ & \textbf{NDS} $\uparrow$ & mATE $\downarrow$ & mASE $\downarrow$ \\
    \midrule
    Lift-Splat-Shoot & 0.446 & 0.486 & 0.742 & 0.489 & 0.336 & 0.416 & 0.781 & 0.510 & 0.367 & 0.340 & 0.764 & 0.631 \\
    PETR        & 0.596 & 0.713 & 0.301 & 0.107 & 0.415 & 0.521 & 0.595 & 0.156 & 0.440 & 0.536 & 0.689 & 0.193 \\
    BEVFormer   & 0.637 & 0.755 & 0.235 & 0.072 & 0.513 & 0.679 & 0.288 & 0.084 & 0.445 & 0.579 & 0.535 & 0.131 \\
    PETRv2      & 0.651 & 0.760 & 0.213 & 0.093 & 0.505 & 0.666 & 0.295 & 0.156 & 0.584 & 0.655 & 0.530 & 0.127 \\
    StreamPETR  & 0.677 & 0.778 & 0.211 & 0.067 & 0.512 & 0.678 & 0.293 & 0.110 & 0.591 & 0.678 &0.489 & 0.102 \\
    BEVNeXt     & 0.681 & 0.781 & 0.206 & 0.069 & 0.527 & 0.686 & 0.291 & 0.112 & 0.603 & 0.694 & 0.447 & 0.103 \\
    GeoBEV      & 0.690 & 0.786 & 0.203 & 0.063 & 0.533 & 0.699 & 0.251 & 0.100 & 0.623 & 0.717 & 0.392 & 0.110 \\
    UVTR        & 0.698 & 0.796 & 0.201 & 0.062 & 0.558 & 0.723 & 0.220 & 0.071 & 0.631 & 0.748 & 0.283 & 0.079 \\
    DETR3D      & 0.701 & 0.782 & 0.289 &  \textbf{0.056} & 0.540 & 0.686 & 0.338 & \textbf{0.063} & 0.677 & 0.764 & 0.320 & 0.069 \\
    \rowcolor{blue!7}
    \textbf{MIC-BEV} & \textbf{0.793} & \textbf{0.844} & \textbf{0.197} & 0.059 & \textbf{0.734} & \textbf{0.809} & \textbf{0.216} & 0.064 & \textbf{0.788} & \textbf{0.839} & \textbf{0.225} & \textbf{0.053} \\
    \bottomrule
  \end{tabular}
  \label{tab:m2i}
\end{table*}

\begin{figure*}
    \centering
    \includegraphics[width=0.98\linewidth]{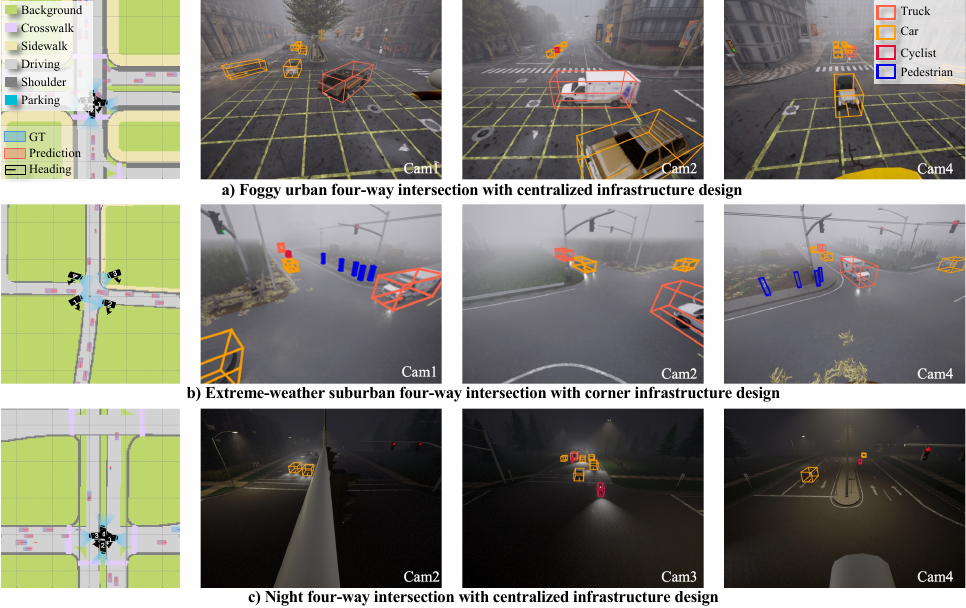}
    \caption{\textbf{Qualitative 3D detection results on the M2I dataset.} MIC-BEV accurately detects multiple object types across diverse urban scenes with varying traffic densities and weather or lighting conditions.}
    \label{fig:m2i}
    \vspace{-0.3cm}
\end{figure*}

\begin{figure*}
    \centering
    \includegraphics[width=0.98\linewidth]{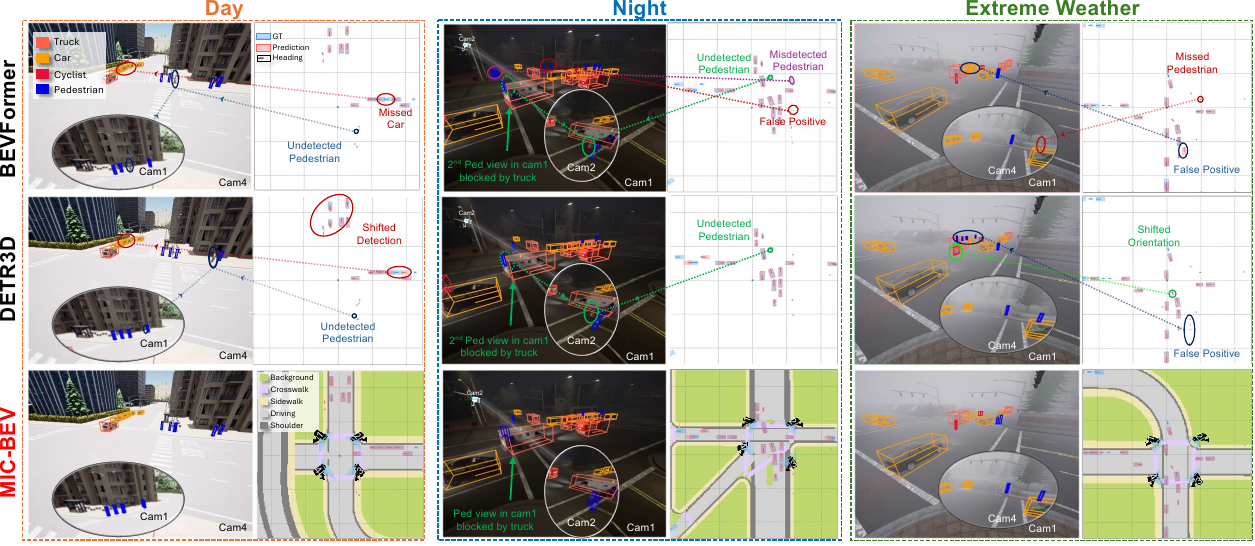}
    \caption{
    \textbf{Qualitative comparison between MIC-BEV and baseline models on the M2I dataset.}
    Compared to BEVFormer and DETR3D, MIC-BEV produces more accurate and consistent detections across day, night, and extreme-weather scenes by leveraging relation-aware multi-view fusion. In the night scene, a pedestrian partially occluded by a truck is missed by the baselines but successfully detected by MIC-BEV.
    }
    \label{fig:comparison}
    \vspace{-0.3cm}
\end{figure*}

\subsection{Results on M2I}
\textbf{Quantitative Evaluation.}
\cref{tab:m2i} summarizes the 3D object detection results on the M2I test set under three evaluation settings: \textit{Normal}, \textit{Robust}, and \textit{Extreme Weather}. MIC-BEV achieves state-of-the-art performance across all settings. Under the Normal condition, MIC-BEV attains 0.793 mAP and 0.844 NDS, outperforming the strongest baseline, DETR3D, by more than 13\% in mAP. The advantage becomes more evident under degraded sensing conditions. In the Robust setting, where one camera is blurred or unavailable, MIC-BEV maintains 0.734 mAP and 0.809 NDS, corresponding to only a 7\% drop from the Normal setting, while competing methods such as DETR3D and BEVFormer degrade by up to 23\%. Under Extreme Weather conditions with heavy rain and dense fog, MIC-BEV continues to achieve the strongest results with 0.788 mAP and 0.839 NDS. These results highlight the effectiveness of MIC-BEV’s relation-enhanced multi-view fusion under diverse simulated camera layouts, environmental conditions, and sensing degradations, enabling robust detection across different road users.

\textbf{Qualitative Evaluation.} 
Qualitative examples of MIC-BEV on M2I are shown in \cref{fig:m2i}. Across simulated foggy urban intersections, extreme-weather suburban scenes, and nighttime settings, MIC-BEV produces accurate 3D  detections for multiple object categories while maintaining consistent BEV segmentation. The segmentation branch provides useful spatial context for road layout and drivable regions, helping distinguish foreground objects from background structures under challenging visibility conditions.

\begin{table*}[t]
  \centering
  \caption{Comparison of 3D object detection performance on the {RoScenes} test set under {Easy} and {Hard} levels.}
  \begin{tabular}{@{}l|ccccc|ccccc|c@{}}
    \toprule
    \multirow{2}{*}{Method} &
    \multicolumn{5}{c|}{\textbf{Easy}} &
    \multicolumn{5}{c|}{\textbf{Hard}} &
    \textbf{Average} \\
    \cmidrule(lr){2-6} \cmidrule(lr){7-11}
    & \textbf{mAP} $\uparrow$ & \textbf{NDS} $\uparrow$ & mATE $\downarrow$ & mASE $\downarrow$ & mAOE $\downarrow$
    & \textbf{mAP} $\uparrow$ & \textbf{NDS} $\uparrow$ & mATE $\downarrow$ & mASE $\downarrow$ & mAOE $\downarrow$ 
    & \textbf{NDS} $\uparrow$ \\
    \midrule
    SOLOFusion   & 0.129 & 0.308 & 0.878 & 0.144 & 0.517 & 0.066 & 0.202 & 0.844 & 0.144 & 1.000 & 0.255 \\
    BEVDet4D     & 0.200 & 0.428 & 0.896 & 0.094 & 0.041 & 0.139 & 0.393 & 0.922 & 0.099 & 0.038 & 0.411 \\
    BEVDet       & 0.299 & 0.506 & 0.742 & 0.079 & 0.042 & 0.184 & 0.445 & 0.754 & 0.087 & 0.043 & 0.476 \\
    StreamPETR   & 0.513 & 0.619 & 0.690 & 0.102 & 0.032 & 0.284 & 0.496 & 0.739 & 0.107 & 0.031 & 0.558 \\
    PETRv2       & 0.587 & 0.674 & 0.590 & 0.090 & 0.032 & 0.414 & 0.580 & 0.633 & 0.100 & 0.029 & 0.627 \\
    BEVFormer    & 0.609 & 0.693 & 0.560 & 0.078 & 0.030 & 0.433 & 0.597 & 0.600 & 0.090 & 0.029 & 0.645 \\
    DETR3D       & 0.644 & 0.722 & 0.501 & 0.067 & 0.031 & 0.471 & 0.633 & 0.508 & 0.080 & 0.028 & 0.678 \\
    RoBEV        & 0.684 & 0.753 & 0.442 & 0.058 & 0.031 & 0.524 & 0.672 & 0.438 & 0.077 & 0.027 & 0.713 \\
    RopeBEV      & 0.721 & 0.786 & 0.435 & \textbf{0.056} & 0.030 & 0.545 & 0.685 & 0.416 & 0.078 & 0.027 & 0.736 \\
    \rowcolor{blue!7}
    \textbf{MIC-BEV} & \textbf{0.743} & \textbf{0.797} & \textbf{0.361} & 0.057 & \textbf{0.027} & \textbf{0.562} & \textbf{0.700} & \textbf{0.384} & \textbf{0.076} & \textbf{0.026} & \textbf{0.749} \\
    \bottomrule
  \end{tabular}
  \label{tab:roscene}
\end{table*}

As shown in \cref{fig:comparison}, MIC-BEV outperforms representative baselines in difficult cases involving occlusion, low illumination, and severe weather. For example, in the night scenario, a pedestrian is partially occluded by a truck in one camera view but remains visible from another camera; MIC-BEV successfully detects the pedestrian by leveraging complementary multi-view observations. Compared with BEVFormer and DETR3D, MIC-BEV reduces missed detections, improves bounding-box alignment, and suppresses false positives through relation-aware multi-view fusion.

\begin{figure*}
    \centering
    \includegraphics[width=0.99\linewidth]{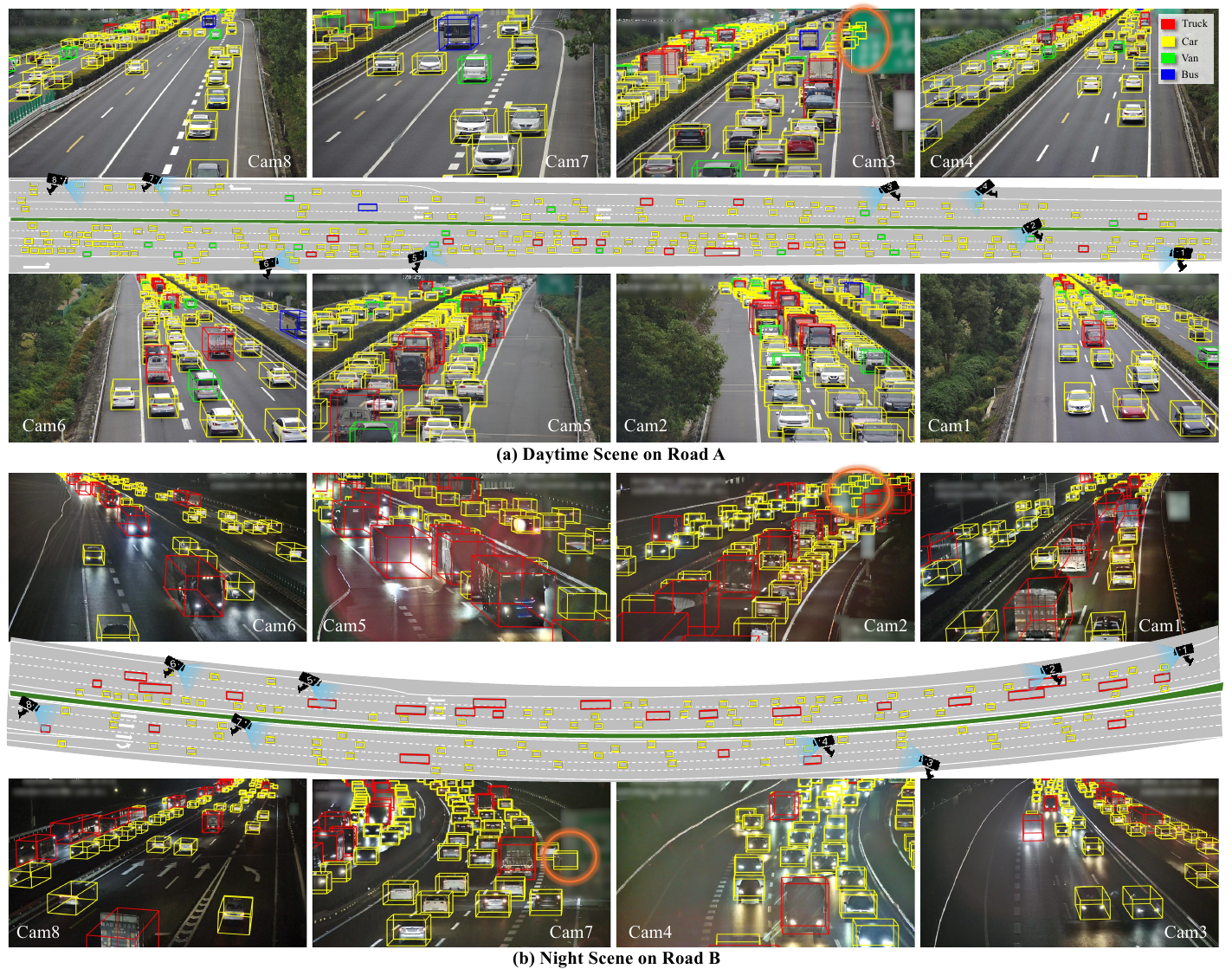}
    \caption{\textbf{Qualitative results of MIC-BEV on RoScenes.}
    Each example shows multi-camera detections and the corresponding fused BEV view over an 800-meter range. MIC-BEV maintains spatially consistent detections across diverse lighting and camera conditions. Orange-circled regions highlight cases where complementary views help recover occluded observations.}
    \label{fig:roscenes}
\end{figure*}

\subsection{Results on RoScenes}
\textbf{Quantitative Evaluation.}
\cref{tab:roscene} presents 3D object detection results on the RoScenes benchmark. MIC-BEV achieves the highest overall performance across both evaluation splits. On the \textit{Easy} set, MIC-BEV attains 0.743 mAP and 0.797 NDS, while on the more challenging \textit{Hard} set, it achieves 0.562 mAP and 0.700 NDS, where issues such as severe occlusion, long-range targets, and limited viewpoint overlap are more prevalent. Compared with the strongest baseline, RopeBEV, MIC-BEV consistently improves mAP and NDS on both splits, reduces mATE by 17\% on the Easy set, and achieves the highest average NDS of 0.749. These results demonstrate that MIC-BEV remains effective in large-scale real-world highway scenes, where reliable long-range localization and robust multi-view fusion are essential for infrastructure-based perception.

\textbf{Qualitative Evaluation.}
\cref{fig:roscenes} shows qualitative results on the RoScenes test set, including synchronized observations from eight roadside cameras and the corresponding fused BEV detections for two representative highway segments. Across both daytime and nighttime scenes, MIC-BEV produces spatially consistent detection results under real-world highway conditions, reliably localizing distant vehicles despite partial occlusion, glare, reflections, and headlight bloom. The orange-circled regions highlight representative cases where objects are partially obscured by roadside structures in one camera view but are successfully recovered through observations from other cameras. These results provide complementary validation of MIC-BEV in real-world highway deployments.

\subsection{Analysis of ReSCA Weights}

\textbf{Quantitative correlation analysis.} 
To examine whether ReSCA weights reflect view-reliability cues, we compute Spearman correlations~\cite{spearman1961proof} between the learned object-center ReSCA weights and several object-camera factors on 100 test scenes from each dataset. Single-camera M2I scenes are excluded because they do not involve cross-view weighting, resulting in 26,510 frames and 3,941,116 object-camera pairs on M2I, and 6,000 frames and 6,298,449 pairs on RoScenes. To avoid confounding viewpoint alignment with projection visibility, viewing-angle deviation is computed only for object-camera pairs with valid projections. For RoScenes, we also use the dataset-provided per-object per-camera occlusion level. The results are summarized in \cref{tab:weight_correlation}.

\begin{table}[t]
\centering
\caption{
Spearman correlations between learned ReSCA weights and object-camera view-reliability factors.
}
\label{tab:weight_correlation}
\setlength{\tabcolsep}{8pt}
\renewcommand{\arraystretch}{1.12}
\begin{tabular}{l|cc}
\toprule
\textbf{View-reliability factor} & \textbf{M2I-Normal} & \textbf{RoScenes} \\
\midrule
Projection validity & 0.894 & 0.818 \\
Viewing angle deviation & -0.310 & -0.514 \\
Distance to image boundary & 0.757 & 0.728 \\
Camera-to-object distance & 0.136 & -0.052 \\
Annotated occlusion level & -- & -0.138 \\
\bottomrule
\end{tabular}
\vspace{-0.3cm}
\end{table}

ReSCA weights show strong positive correlations with projection validity and distance to the image boundary on both datasets, indicating that visible and less boundary-truncated observations receive higher weights. The negative correlations with viewing-angle deviation on both datasets further show that ReSCA favors better-aligned views. In contrast, camera-to-object distance has only a weak positive correlation on M2I and a negligible correlation on RoScenes. This weak and inconsistent relationship suggests that ReSCA does not simply select the nearest camera but instead relies more on projection validity, view coverage, and viewing geometry. This is expected in infrastructure-camera settings, where a closer camera may still provide an oblique or boundary-truncated view, while a farther camera may offer better image coverage.

On RoScenes, the annotated occlusion level shows a weak negative correlation with ReSCA weights. This trend is directionally consistent with more occluded observations receiving lower weights, but its small magnitude suggests that occlusion is not the dominant weighting factor. Overall, these results show that ReSCA weights are aligned with meaningful view-reliability cues across both synthetic and real-world datasets.

\begin{figure*}
    \centering
    \includegraphics[width=0.96\linewidth]{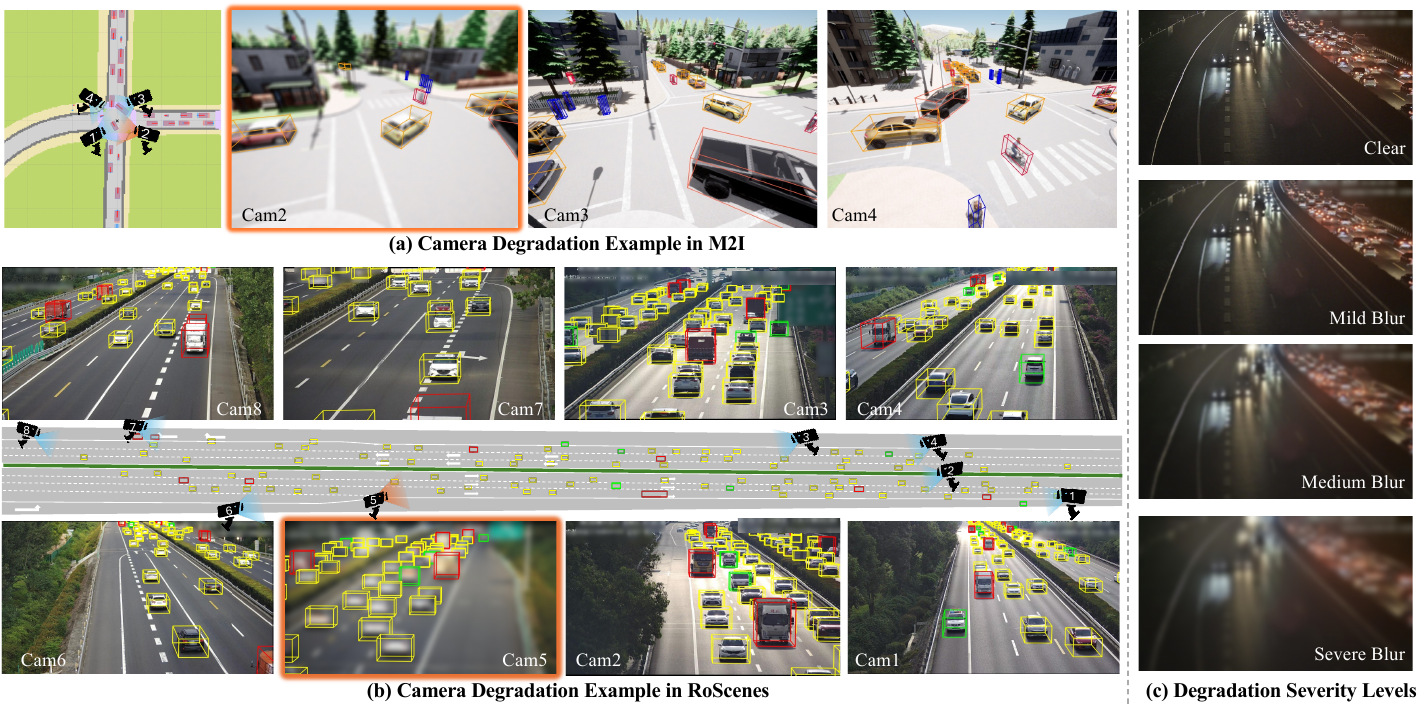}
     \caption{
    \textbf{Examples of visual degradation in M2I and RoScenes.}
    Orange-highlighted views indicate degraded cameras affected by blur, while other cameras provide clearer complementary observations. (a) and (b) show representative degraded-camera scenarios in M2I and RoScenes, while (c) illustrates the mild, medium, and severe blur levels used in our robustness analysis.
    }
    \label{fig:camera_degradation}
\end{figure*}

\textbf{Weight drop under camera degradation.}
We further investigate whether ReSCA reduces reliance on degraded views. The analysis is conducted on 100 four-camera M2I test scenarios covering 26,753 frames. We focus on four-camera scenes because they provide sufficient cross-view redundancy, allowing the model to rely on alternative views when one camera is degraded. For each sampled scenario, we first run clean inference with camera masking disabled, and then rerun inference on the same frames after deterministically corrupting one camera using Gaussian blur or full camera masking. We perturb one camera at a time to isolate the effect of an individual degraded sensor. For each valid object whose center is projected into the corrupted view, we extract the ReSCA weight assigned to that camera at the object-center BEV cell. We then compute the relative weight drop between clean and corrupted inference and average it over all valid objects and frames. Representative detection examples and visualizations for different degradation severity levels are shown in \cref{fig:camera_degradation}.

As shown in \cref{tab:weight_degradation_response}, ReSCA progressively downweights the corrupted view as degradation becomes stronger. The average weight drop increases from 6.3\% under mild blur to 14.2\% under severe blur, while full camera masking produces the largest reduction of 20.8\%. This behavior demonstrates that ReSCA performs adaptive rather than uniform camera fusion. The weight reduction is also gradual rather than binary, indicating that mildly degraded views are still partially used when they contain useful information. Overall, these results provide quantitative evidence that ReSCA learns reliability-aware view weighting under sensor degradation, reducing the influence of unreliable cameras while retaining useful information from partially degraded views.

\begin{table}[t]
\centering
\caption{
Relative ReSCA weight change under visual degradation on the four-camera M2I-Normal test set.
}
\label{tab:weight_degradation_response}
\setlength{\tabcolsep}{5pt}
\renewcommand{\arraystretch}{1.12}
\begin{tabular}{l|ccc}
\toprule
\textbf{Condition} & \textbf{Weight Drop}  & \textbf{mAP} $\uparrow$ & \textbf{NDS} $\uparrow$ \\
\midrule
Clean view                            & --     & 0.811 & 0.854 \\
Mild blur ($\sigma=10$, kernel=11)    & 6.3\%  & 0.789 & 0.840 \\
Medium blur ($\sigma=10$, kernel=61)  & 12.6\% & 0.767 & 0.828 \\
Severe blur ($\sigma=20$, kernel=121) & 14.2\% & 0.748 & 0.817  \\
Camera masking                        & 20.8\% & 0.747 & 0.816  \\
\bottomrule
\end{tabular}
\vspace{-0.3cm}
\end{table}

\begin{figure*}
    \centering
    \includegraphics[width=\linewidth]{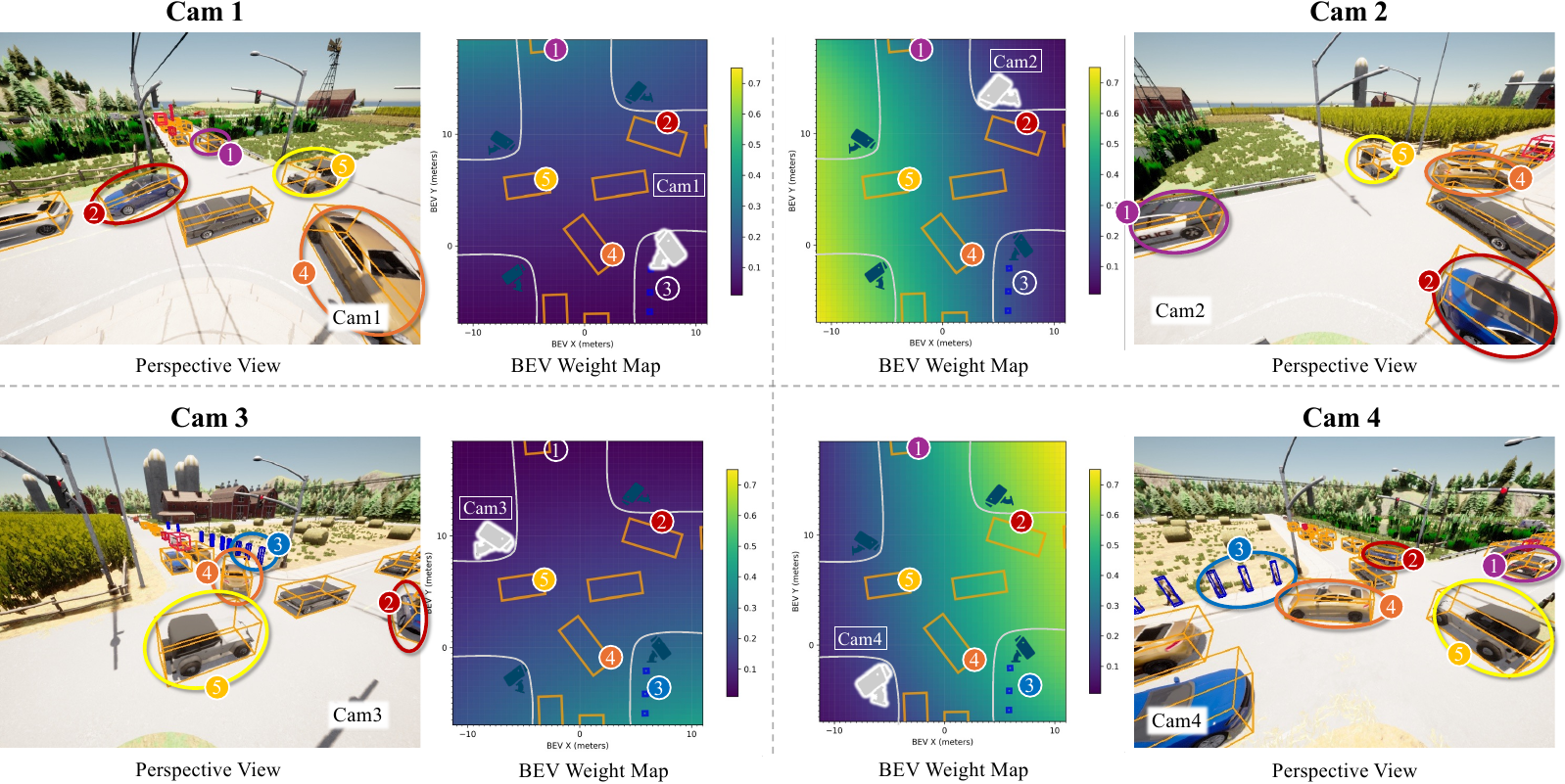}
    \caption{
    \textbf{Camera-wise importance weights for BEV grid cells at an intersection.}
    For each camera (Cam1-Cam4), the perspective view and corresponding BEV importance weight map are shown. Numbered objects (\ding{172}-\ding{176}) are representative examples discussed in the text. Colored markers in the BEV weight map denote objects visible in the corresponding camera, while uncolored markers denote occluded or out-of-view objects. The result shows that each BEV cell adaptively assigns higher weights to the camera views that provide the most informative observations based on visibility and viewing geometry.
    }
    \label{fig:attention}
\end{figure*}

\textbf{Qualitative analysis.} 
\cref{fig:attention} illustrates the learned camera-wise importance weights assigned by the GNN in MIC-BEV across four cameras monitoring an intersection. The model assigns higher weights to cameras where objects are clearly visible, centered in the field of view, and minimally distorted.

For example, \textit{Object~4} is visible in all four cameras but appears heavily elongated in Cam1 due to an oblique viewing angle. Correspondingly, MIC-BEV assigns a lower weight to Cam1 and emphasizes Cam2 and Cam4, which observe the object from less distorted, more front-facing perspectives. Similarly, \textit{Object~2} is captured in all views but lies near the image boundary in Cam2 and Cam3, where partial occlusion occurs. The model appropriately prioritizes Cam4, where the object is well-centered and clearly observed. \textit{Object~1} appears in Cam1, Cam2, and Cam4, but these views capture it at a distance or near the periphery. Hence, the GNN assigns roughly uniform weights. \textit{Object~3} is visible only in Cam3 and Cam4, both offering direct, minimally occluded views. MIC-BEV assigns balanced and high weights to both cameras. Finally, \textit{Object~5}, seen across all cameras, receives the highest weight from Cam2, which observes the object near the center with the clearest geometric fidelity, consistent with ReSCA's tendency to favor informative and geometrically reliable views.

\subsection{Ablation Studies}

\textbf{Component Ablation.}
We perform a component ablation on the M2I test set to evaluate the contribution of each key module in MIC-BEV, including camera masking, the auxiliary BEV segmentation head, and the ReSCA module.
As shown in \cref{tab:ablation}, removing any component leads to a consistent performance drop under both the Normal and Robust settings.

\begin{table}[ht]
  \centering
  \caption{Ablation study of MIC-BEV components on M2I.
  \textit{Cam. Masking}: random camera masking; 
  \textit{BEV Seg.}: the auxiliary BEV segmentation head; 
  \textit{ReSCA}: Relation-enhanced spatial cross-attention module}
\begin{tabular}{@{}ccc|cc|cc@{}}
    \toprule
    \multirow{2}{*}{\makecell{Cam.\\Masking}} &
    \multirow{2}{*}{\makecell{BEV\\Seg.}} &
    \multirow{2}{*}{\makecell{ReSCA}} &
    \multicolumn{2}{c|}{\textbf{M2I-Normal}} &
    \multicolumn{2}{c}{\textbf{M2I-Robust}} \\
    \cmidrule(lr){4-5}\cmidrule(l){6-7}
     & & & mAP$\uparrow$ & NDS$\uparrow$ & mAP$\uparrow$ & NDS$\uparrow$ \\
    \midrule
    \xmark & \xmark & \xmark & 0.637 & 0.755 & 0.513 & 0.679 \\
    \cmark & \xmark & \xmark & 0.651 & 0.767 & 0.624 & 0.701 \\
    \cmark & \cmark & \xmark & 0.717 & 0.792 & 0.669 & 0.737 \\
    \rowcolor{blue!7}
    \cmark & \cmark & \cmark & \textbf{0.793} & \textbf{0.844} & 
    \textbf{0.734} & \textbf{0.809} \\
    \bottomrule
  \end{tabular}
  \label{tab:ablation}
\end{table}

Enabling random camera masking improves robustness by encouraging the model to rely on complementary camera views.
The auxiliary BEV segmentation head further enhances spatial reasoning, providing geometric priors that guide detection.
Finally, by studying the effect of the ReSCA module, we replace it with a regular spatial cross-attention layer without graph-based weighting and observe a clear reduction in both mAP and NDS.
This confirms that the GNN weighting mechanism in ReSCA effectively learns view-dependent importance for multi-camera fusion, leading to stronger performance.

\begin{table}[t]
\centering
\caption{Influence of incorporating camera features, distance, and angles in the GNN-based multi-view fusion module.}
\setlength{\tabcolsep}{8pt}
\begin{tabular}{c c c | c c}
\toprule
\multirow{2}{*}{\textbf{Cam. Feature}} &
\multirow{2}{*}{\textbf{Distances}} &
\multirow{2}{*}{\textbf{Angles}} &
\multicolumn{2}{c}{\textbf{M2I-Normal}} \\
\cmidrule(l){4-5}
 & & & \textbf{mAP$\uparrow$} & \textbf{NDS$\uparrow$} \\
\midrule
\xmark & \xmark & \xmark & 0.717 & 0.792 \\
\xmark & \cmark & \cmark & 0.738 & 0.794 \\
\cmark & \xmark & \xmark & 0.733 & 0.789 \\
\cmark & \cmark & \xmark & 0.756 & 0.806 \\
\rowcolor{blue!7}
\cmark & \cmark & \cmark & \textbf{0.793} & \textbf{0.844} \\
\bottomrule
\end{tabular}
\label{tab:gnn_ablation}
\vspace{-0.2cm}
\end{table}

\textbf{Influence of Relation Encoding.}
\cref{tab:gnn_ablation} evaluates the effect of different inputs on the GNN-based multi-view fusion module. Using only the baseline BEV representation yields 0.717 mAP and 0.792 NDS. Adding geometric cues alone, including distances and angles, improves mAP to 0.738, indicating that explicit camera-BEV geometry provides useful guidance for view selection. Camera features alone also improve performance to 0.733 mAP, but the gain is smaller, suggesting that appearance cues are beneficial but insufficient without geometric context. Combining camera features with distance cues further improves performance to 0.756 mAP and 0.806 NDS. The full model, which incorporates camera features, distances, and angles, achieves the best result with 0.793 mAP and 0.844 NDS. This confirms that ReSCA benefits from jointly modeling visual reliability and geometric relationships during multi-camera BEV fusion.

\begin{figure*}
    \centering
    \includegraphics[width=0.99\linewidth]{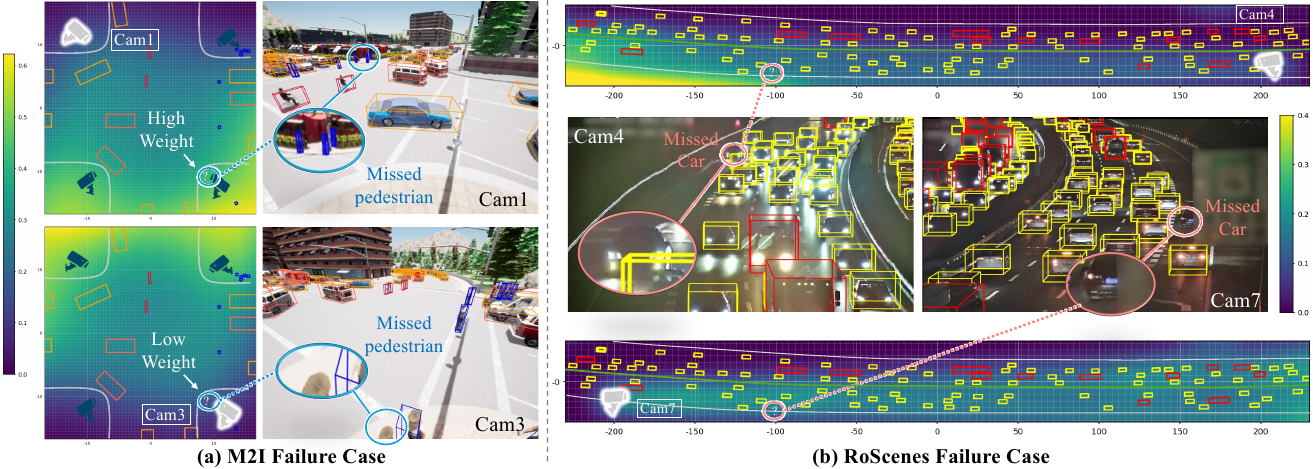}
    \caption{
    \textbf{Representative failure cases of MIC-BEV.} ReSCA weight maps and corresponding camera views illustrate two representative failure modes: suboptimal view weighting, where useful local evidence is underweighted (M2I), and information-limited detection, where all available views provide insufficient evidence despite reasonable weighting (RoScenes).
    }
    \label{fig:failure}
    \vspace{-0.2cm}
\end{figure*}

\textbf{Robustness to sensor degradation.} To evaluate robustness under realistic infrastructure-camera failures, we conduct controlled degradation experiments on the test sets of M2I and RoScenes. For visual degradation and camera masking, we randomly perturb one camera at a time in multi-camera scenarios while keeping the remaining cameras unchanged. The blur settings follow \cref{tab:weight_degradation_response}, and representative examples are shown in \cref{fig:camera_degradation}. Calibration perturbations are grouped into three levels: mild ($|\Delta r|\le1^\circ$, $|\Delta t|\le0.1$\,m), medium ($1^\circ<|\Delta r|\le3^\circ$, $0.1<|\Delta t|\le0.3$\,m), and severe ($3^\circ<|\Delta r|\le5^\circ$, $0.3<|\Delta t|\le0.5$\,m). Rotation ($\Delta r$) and translation ($\Delta t$) perturbations are randomly sampled within the specified ranges and applied to the selected camera extrinsics.

\begin{table}[t]
\centering
\caption{Robustness evaluation under visual degradation, camera masking, and calibration noise on M2I-Normal and RoScenes. \textit{Base} denotes training on clean data, while \textit{Aug.} denotes training with degradation data augmentation. Results are reported using NDS.}
\label{tab:robustness_analysis}
\setlength{\tabcolsep}{4pt}
\renewcommand{\arraystretch}{1.12}
\begin{tabular}{ll|cc|cc}
\toprule
\multirow{2}{*}{\textbf{Category}}
& \multirow{2}{*}{\textbf{Condition}}
& \multicolumn{2}{c|}{\textbf{M2I-Normal}}
& \multicolumn{2}{c}{\textbf{RoScenes}} \\
\cmidrule(lr){3-4}
\cmidrule(lr){5-6}
& & Base & Aug. & Base & Aug. \\
\midrule

Clean
& Clear view
& \textbf{0.850}
& 0.844
& 0.745
& \textbf{0.749} \\

\midrule
\multirow{4}{*}{Visual degradation}
& Mild blur
& 0.789
& \textbf{0.829}
& 0.731
& \textbf{0.744} \\

& Medium blur
& 0.773
& \textbf{0.814}
& 0.710
& \textbf{0.735} \\

& Severe blur
& 0.772
& \textbf{0.797}
& 0.709
& \textbf{0.724} \\

& Camera masking
& 0.769
& \textbf{0.789}
& 0.706
& \textbf{0.717} \\

\midrule
\multirow{3}{*}{Calibration noise}
& Mild
& 0.819
& \textbf{0.828}
& 0.678
& \textbf{0.699} \\

& Medium
& 0.749
& \textbf{0.777}
& 0.656
& \textbf{0.693} \\

& Severe
& 0.719
& \textbf{0.737}
& 0.651
& \textbf{0.692} \\

\bottomrule
\end{tabular}
\vspace{-0.5cm}
\end{table}

As summarized in \cref{tab:robustness_analysis}, degradation-augmented training consistently improves robustness under visual corruption and camera masking. On M2I, the augmented model incurs only a small reduction on clean data from 0.850 to 0.844, but consistently outperforms the clean-trained base model under all degraded visual settings, for example, improving NDS from 0.769 to 0.789 under camera masking. Similar improvements are observed on RoScenes across all visual degradation levels. These results indicate that degradation data augmentation encourages MIC-BEV to better handle degraded observations and exploit complementary views.

Calibration errors remain particularly challenging on both datasets. Although degradation-augmented training consistently improves robustness over the clean-trained base model, the performance still decreases from 0.844 to 0.737 on M2I under severe calibration perturbation. A similar trend is observed on RoScenes, where NDS decreases from 0.749 to 0.692 under severe calibration noise. Unlike blur, which primarily affects local appearance, calibration errors directly perturb image-to-BEV projection and cross-view spatial alignment, degrading the geometric consistency required for multi-view fusion. The larger performance drop in RoScenes, even under mild calibration noise, is consistent with its long-range highway setting, where small extrinsic errors can produce substantial projection shifts for distant objects. Overall, MIC-BEV is robust to appearance degradation and partial camera failure, while accurate camera calibration remains essential for reliable long-range infrastructure perception.

\section{Discussions}
\subsection{Failure Case Diagnostics}
We further examine representative failure cases to illustrate the limitations of ReSCA under challenging conditions. As shown in \cref{fig:failure}, the examples reveal two failure modes: suboptimal view weighting, where the learned weights do not fully align with the most informative local evidence, and information-limited detection, where the weighting is reasonable but all available views provide insufficient evidence.

In the first example from M2I, shown in \cref{fig:failure}(a), the missed pedestrian is weakly observable in both views. Cam1 provides a more geometrically favorable observation, but the pedestrian appears extremely small; Cam3 contains stronger local evidence, but the pedestrian lies near the image boundary. Consistent with \cref{tab:weight_correlation}, ReSCA assigns a higher weight to Cam1 and downweights the boundary-region observation in Cam3. In this case, the weight assignment is suboptimal because the closer Cam3 view provides stronger local evidence despite its boundary placement. This failure suggests that ReSCA may occasionally overemphasize geometric reliability when local appearance cues are weak, leading to missed detections for small boundary-region objects.

The second example from RoScenes, shown in \cref{fig:failure}(b), illustrates an information-limited case under long-range nighttime perception. The target vehicle is visible in both camera views, but neither observation is satisfactory. In Cam4, the vehicle is distant and partially occluded by a preceding vehicle, while in Cam7, it is partially occluded by a roadside sign. ReSCA assigns comparable weights to both views because neither provides a clearly superior observation. However, the vehicle is still missed because both views contain insufficient visual evidence for reliable detection. This case indicates that even reasonable weight assignment cannot compensate when all available observations are similarly limited.

\subsection{Computational Efficiency and Deployment Analysis}

\textbf{Runtime Comparison with Existing Methods.}
We profile MIC-BEV and all baselines using their official implementations under a unified evaluation protocol: the M2I-Normal test split, batch size 1, a fixed four-camera input configuration, and a single NVIDIA L40S GPU. As summarized in \Cref{tab:cost}, MIC-BEV achieves the best detection performance, reaching 0.793 mAP, which improves upon the strongest baseline, DETR3D, by 9.2 mAP points. This accuracy improvement comes with increased computational cost, achieving 8.82 FPS with 3.68 GB GPU memory. Nevertheless, its memory consumption remains comparable to other Transformer-based BEV detectors such as BEVFormer (3.64 GB) and PETRv2 (3.28 GB), and is substantially lower than GeoBEV (7.51 GB) and UVTR (6.10 GB). Overall, these results show that MIC-BEV achieves state-of-the-art detection performance while maintaining a moderate inference cost for multi-camera infrastructure perception.

\begin{table}[t]
\centering
\caption{
Performance and inference cost of detection models on a single NVIDIA L40S GPU with batch size 1.
}
\label{tab:cost}
\setlength{\tabcolsep}{5pt}
\begin{tabular}{l|ccc|c}
\toprule
\textbf{Method} & \textbf{Params (M)} & \textbf{Memory (GB)} & \textbf{FPS} & \textbf{mAP $\uparrow$}\\
\midrule
LSS              & 21.51 & 2.05 & 17.02 & 0.446 \\
PETR             & 37.26 & 2.77 & 16.51 & 0.596\\
BEVFormer        & 68.67 & 3.64 & 9.97 & 0.637\\
PETRv2           & 61.53 & 3.28 & 12.94 & 0.651 \\
StreamPETR       & 37.26 & 2.65 & 14.05 & 0.677 \\
BEVNeXt          & 65.43 & 2.36 & 7.66  & 0.681 \\
GeoBEV           & 75.75 & 7.51 & 12.88 & 0.690 \\
UVTR             & 38.96 & 6.10 & 13.01 & 0.698 \\
DETR3D           & 53.64 & 2.48 & 16.20 & 0.701 \\  
\rowcolor{blue!7}
\textbf{MIC-BEV }         & 73.50 & 3.68 & 8.82  & \textbf{0.793} \\
\bottomrule
\end{tabular}
\end{table}

\textbf{Scalability Analysis.}
We analyze the deployment scalability of MIC-BEV with respect to BEV resolution and camera count on the M2I-Normal test set, as summarized in \cref{tab:scaling}.

For BEV-resolution scalability, we train separate MIC-BEV models with different BEV grid sizes while fixing the maximum camera capacity to four cameras. Increasing the resolution consistently improves accuracy, with mAP increasing from 0.765 at $100\times100$ to 0.818 at $300\times300$. However, this improvement comes with a higher computational cost: the $300\times300$ setting achieves the best accuracy but reduces throughput to 3.02 FPS and increases memory usage to 7.62 GB. In contrast, the $100\times100$ setting provides an efficiency-oriented operating point, reaching 14.24 FPS with 2.59 GB memory while retaining 0.765 mAP and 0.815 NDS. Compared with the default $200\times200$ setting, it improves throughput by 61.5\% and reduces memory usage by 29.6\%, with a moderate 2.8 mAP decrease. These results show that MIC-BEV can be configured for either runtime-sensitive or accuracy-oriented deployment according to system requirements.

\begin{table}[t]
\centering
\caption{
Scalability analysis on M2I. The BEV-resolution experiment fixes the max input to 4 cam, while the camera-count experiment fixes the BEV resolution at $200\times200$ and varies the max camera capacity. $\dagger$ denotes the default setting.
}
\label{tab:scaling}
\setlength{\tabcolsep}{2.5pt}
\begin{tabular}{ll|ccc|cc}
\toprule
\textbf{Category} & \textbf{Setting} & \textbf{Params} & \textbf{Memory} & \textbf{FPS} & \textbf{mAP $\uparrow$} & \textbf{NDS $\uparrow$} \\
\midrule

\multirow{3}{*}{\textbf{BEV Resolution}}
& 100$\times$100 & 65.793 & 2.59 & 14.24 & 0.765 & 0.815 \\
& 200$\times$200$^\dagger$ & 73.498 & 3.68 & 8.82 & 0.793 & 0.844 \\
& 300$\times$300& 86.324 & 7.62 & 3.02 & \textbf{0.818} & \textbf{0.859} \\
\midrule
\multirow{4}{*}{\textbf{Camera Count}}
& 1 cam & 73.494 & 2.76 & 15.09 & 0.451 & 0.623 \\
& 2 cams & 73.496 & 3.05 & 12.92 & 0.625 & 0.727 \\
& 3 cams & 73.497 & 3.43 & 10.10 & 0.726 & 0.790 \\
& 4 cams$^\dagger$ & 73.498 & 3.68 & 8.82 & \textbf{0.793} & \textbf{0.844} \\
\bottomrule
\end{tabular}
\end{table}

For camera-count scalability, we train MIC-BEV models with different maximum camera capacities, ranging from one to four cameras, under the default $200\times200$ BEV resolution. For each setting, scenarios with fewer cameras are padded, while scenarios with more cameras are evaluated by randomly sampling from the available cameras. This protocol is used to analyze the effect of maximum camera capacity and available scene coverage. The results show that increasing the maximum camera capacity consistently improves detection performance, from 0.451 mAP and 0.623 NDS with one camera to 0.793 mAP and 0.844 NDS with four cameras, while reducing inference throughput from 15.09 FPS to 8.82 FPS and increasing memory from 2.76 GB to 3.68 GB. This result indicates that additional cameras provide broader spatial coverage and stronger cross-view complementarity, but also introduce higher computational cost. Together with the BEV-resolution study, these results demonstrate configurable operating points for balancing accuracy, memory, and inference throughput under different deployment requirements.

%%%%%%%%%%%%%%%%%%%%%%%%%%%%%%%%%%%%%%%%%%%%%%%%%%%%%%%%%%%%%%%%%%%%%%%%%%%%%%%%%%%%%%
\section{Conclusions}
We propose \textbf{MIC-BEV}, a Transformer-based framework for multi-camera infrastructure perception, together with \textbf{M2I}, a comprehensive synthetic perception dataset and benchmark covering diverse road layouts, camera configurations, illumination conditions, and adverse weather. MIC-BEV introduces a \textbf{camera-BEV spatial relation-aware attention} mechanism that models geometric relations between cameras and the BEV grid through a graph neural network, enabling geometry-aware adaptive multi-view fusion under heterogeneous camera configurations. It further incorporates BEV segmentation and degradation-augmented training to improve spatial reasoning and robustness against occlusion, sensor degradation, camera failures, and calibration perturbations.

Extensive experiments demonstrate state-of-the-art performance on both synthetic and real-world datasets. On M2I, MIC-BEV achieves strong detection performance across diverse simulated infrastructure layouts, camera configurations, illumination conditions, and adverse weather. On RoScenes, it provides complementary real-world validation in large-scale highway environments, demonstrating robust performance under real infrastructure-camera geometry and long-range vehicle detection. MIC-BEV also improves robustness to camera degradation, temporary sensor failures, and calibration perturbations through degradation-augmented training and adaptive fusion, although accurate extrinsic calibration remains important for long-range perception. Weight analysis for the fusion module further shows that the learned fusion weights strongly correlate with geometric visibility cues, including projection validity, image-boundary distance, and viewing angle.

Despite these promising results, several limitations remain. First, detecting small objects under weak or boundary-region observations remains challenging. Second, although M2I provides diverse simulated environments and RoScenes offers complementary real-world validation, the current real-world evaluation is limited to highway scenes and vehicle detection. Future work will extend MIC-BEV to broader infrastructure perception tasks, including multi-object tracking and trajectory forecasting, to better capture dynamic road-user interactions. Additional directions include incorporating complementary sensing modalities such as radar, improving computational efficiency through model compression and optimized operators, and validating the framework across more diverse real-world infrastructure deployments, including dense urban intersections, vulnerable road users, and adverse weather conditions.

\bibliographystyle{IEEEtran}
\bibliography{refs}

\begin{IEEEbiography}[{\includegraphics[width=1in,height=1.25in,clip,keepaspectratio]{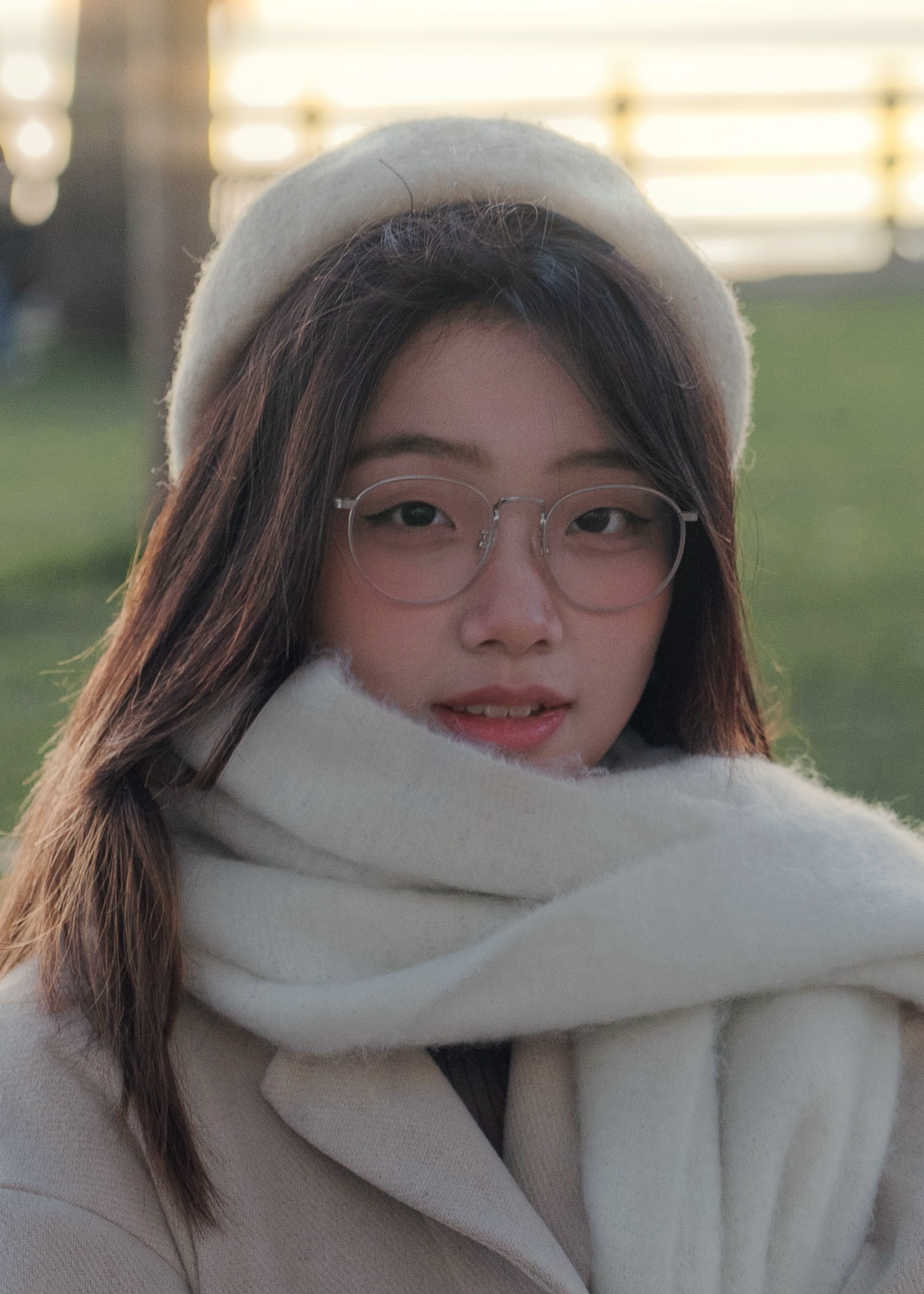}}]{Yun Zhang} (Student Member, IEEE) is a Computer Science Ph.D. student at the University of California, Los Angeles, where she is a member of the Mobility Lab and Scalable Analytics Institute (ScAi) Lab, and is an Amazon Trainium Fellowship awardee. She received her B.S. degrees in Computational Mathematics and Statistics and Data Science from UCLA in 2025. Her research interests lie in computer vision, autonomous systems, and embodied intelligence, with a focus on multi-camera perception, multi-agent cooperation, and vision-language-action modeling for robots and intelligent transportation.
\end{IEEEbiography}

\begin{IEEEbiography}[{\includegraphics[width=1in,height=1.25in,clip,keepaspectratio]{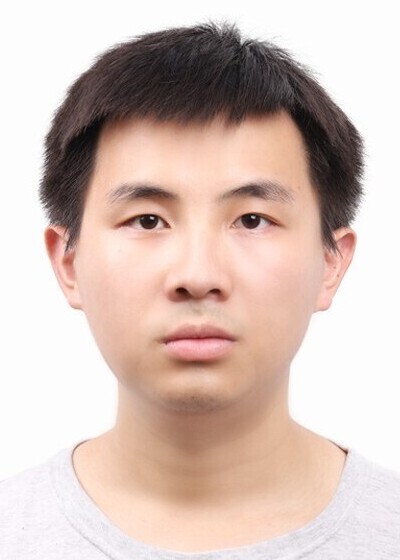}}]{Zhaoliang Zheng} (Student Member, IEEE)
received the B.E. degree in Process Equipment and Control Engineering from Dalian University of Technology, China, in 2017, and the M.S. degree in Mechanical and Aerospace Engineering from the University of California, San Diego, in 2019. He is currently pursuing a Ph.D. degree in Electrical and Computer Engineering at the UCLA Mobility Lab, University of California, Los Angeles. His research interests include robotics, embedded systems, sensing and planning for autonomous driving, and V2X cooperative perception. 
\end{IEEEbiography}

\begin{IEEEbiography}[{\includegraphics[width=1in,height=1.25in,clip,keepaspectratio]{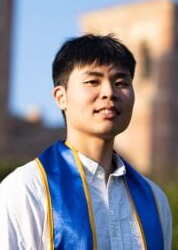}}]{Johnson Liu} is a Research and Development Engineer at the UCLA Mobility Lab. He earned his M.S. in Electrical and Computer Engineering from the University of California, Los Angeles. in 2025 and his B.S. in Computer Science from SUNY Oswego in 2023. His research focuses on intelligent systems, embedded computing, and real-time perception and control for autonomous vehicles and connected infrastructure.
\end{IEEEbiography}

\begin{IEEEbiography}[{\includegraphics[width=1in,height=1.25in,clip,keepaspectratio]{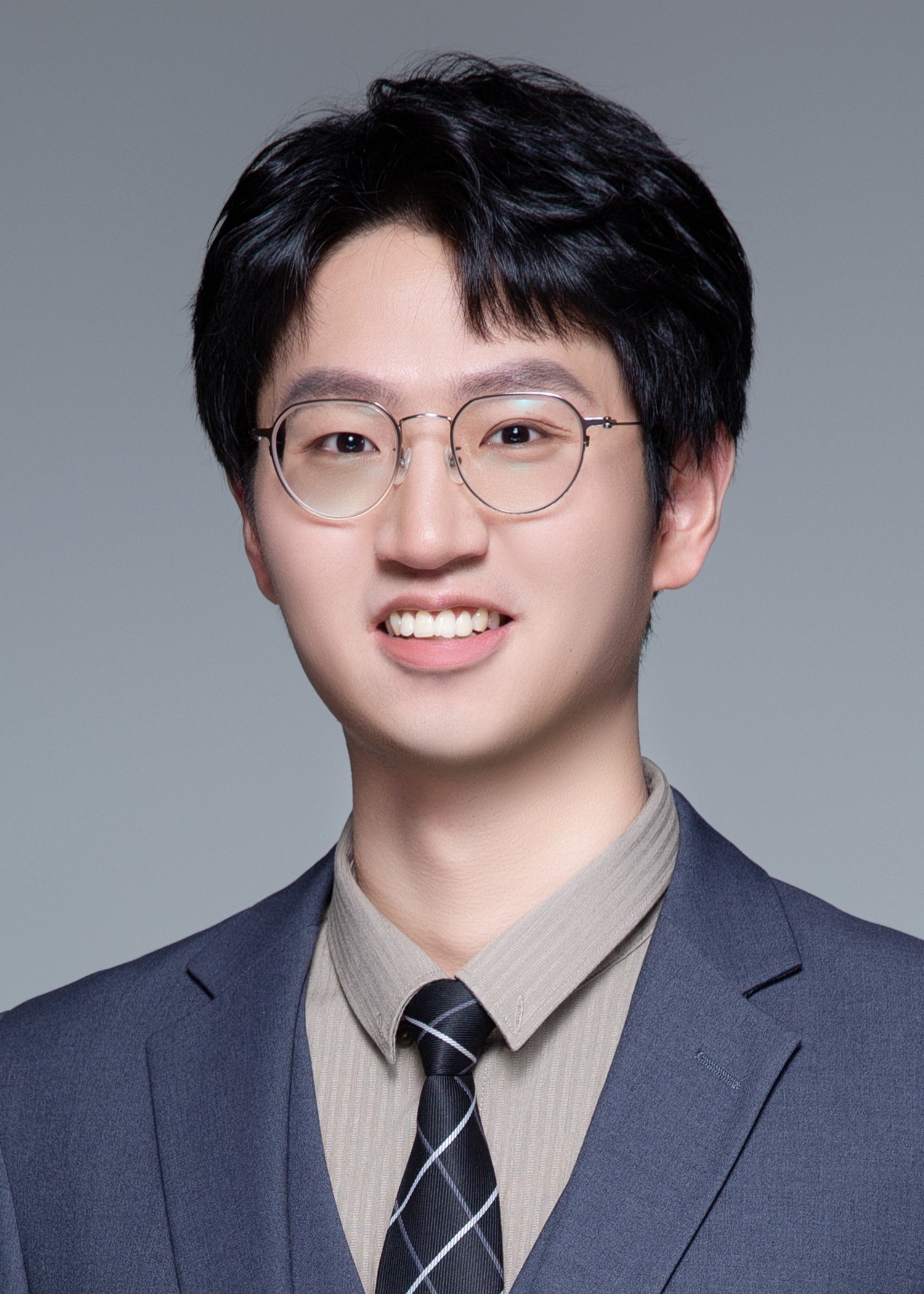}}]{Zhiyu Huang} (Member, IEEE)
 is a Postdoctoral Scholar at the University of California, Los Angeles. He received the Ph.D. degree from Nanyang Technological University (NTU), Singapore, in 2024, and the B.Eng. degree from Chongqing University, China, in 2019. His research focuses on algorithmic development for embodied AI agents, particularly in areas of perception, prediction, reasoning, decision-making, and simulation for robotics and autonomous vehicles. He has authored numerous papers in leading journals and conferences and received recognitions such as the NTU MAE Best Ph.D. Thesis Award and IEEE ITSS Best Dissertation Award Finalist. He also serves as an Associate Editor of the IEEE Open Journal of Intelligent Transportation Systems.
\end{IEEEbiography}

\begin{IEEEbiography}[{\includegraphics[width=1in,height=1.25in,clip,keepaspectratio]{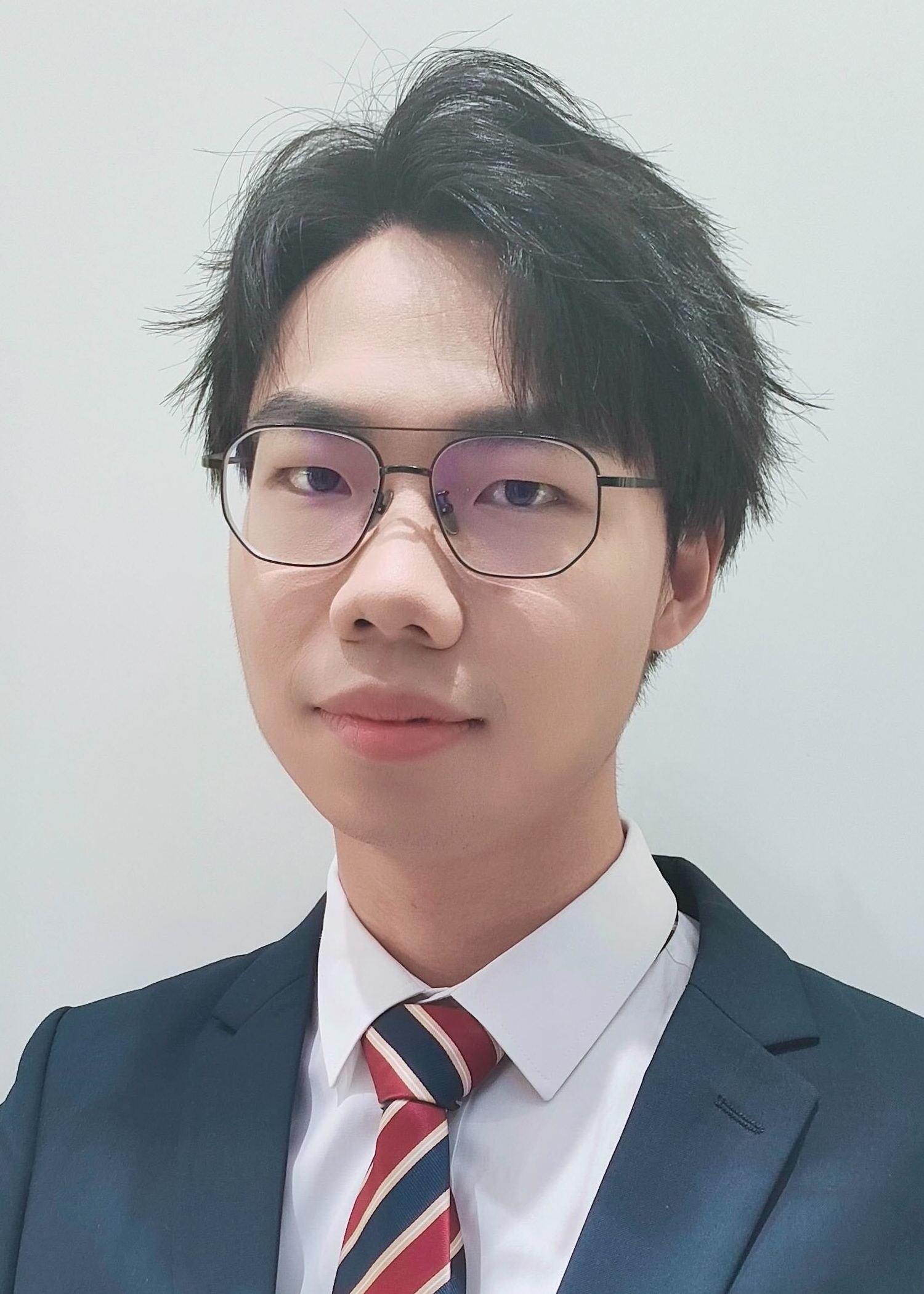}}]{Zewei Zhou} (Student Member, IEEE) is a Ph.D. candidate at the University of California, Los Angeles and an Amazon AI PhD Fellowship awardee. He received the M.S. degree from Tongji University, Shanghai, China, in 2023, and the B.Eng. degree from Chongqing University, Chongqing, China, in 2020. His research interests include embodied AI, autonomous driving, and cooperative automation, with a focus on scene understanding and reasoning for perception, prediction, planning, and simulation tasks.
\end{IEEEbiography}

\begin{IEEEbiography}[{\includegraphics[width=1in,height=1.25in,clip,keepaspectratio]{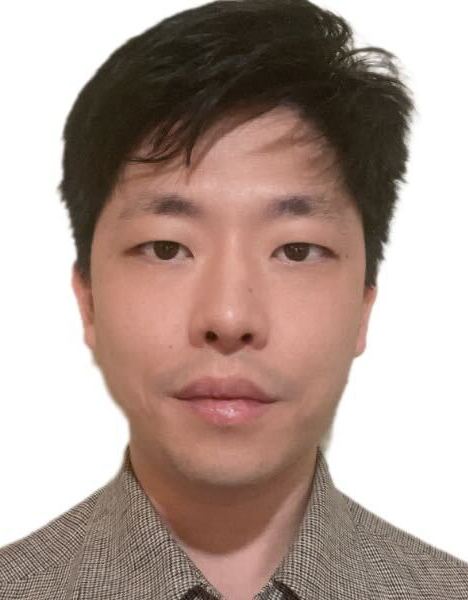}}]{Zonglin Meng} (Member, IEEE) is a postdoctoral researcher with the UCLA Mobility Lab, Los Angeles, CA, USA. His research focuses on autonomous driving, particularly multimodal sensor fusion, perception, and multi-agent coordination for robust, scalable autonomy. He received the B.S. degree from the University of Wisconsin–Madison in 2020 and the Ph.D. degree from the University of California, Los Angeles, in 2024. During his doctoral studies at UCLA, he worked on learning-based perception and cooperative driving strategies across heterogeneous sensors and platforms.
\end{IEEEbiography}

\begin{IEEEbiography}[{\includegraphics[width=1in,height=1.25in,clip,keepaspectratio]{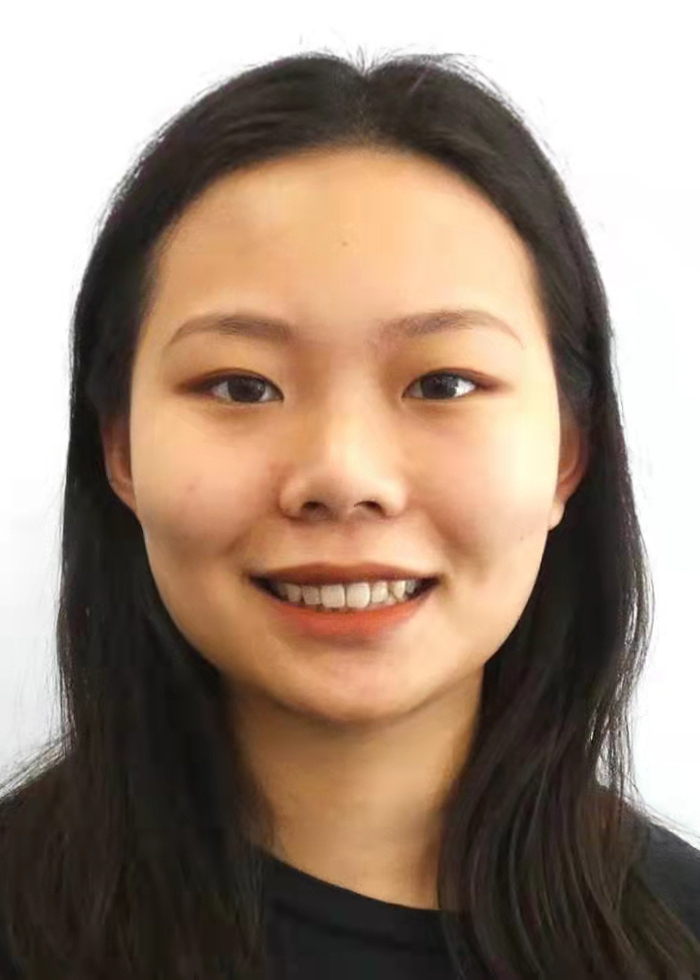}}]{Tianhui Cai} (Student Member, IEEE) is a Ph.D. student at the University of California, Los Angeles. She received the M.S. degree from Carnegie Mellon University in 2023 and the B.S. degree from the University of Illinois at Urbana-Champaign in 2022. Her research focuses on visual representation learning and generative modeling, with applications in autonomous driving, particularly in the areas of perception, planning, and 2D/3D scene reconstruction.
\end{IEEEbiography}

\begin{IEEEbiography}[{\includegraphics[width=1in,height=1.25in,clip,keepaspectratio]{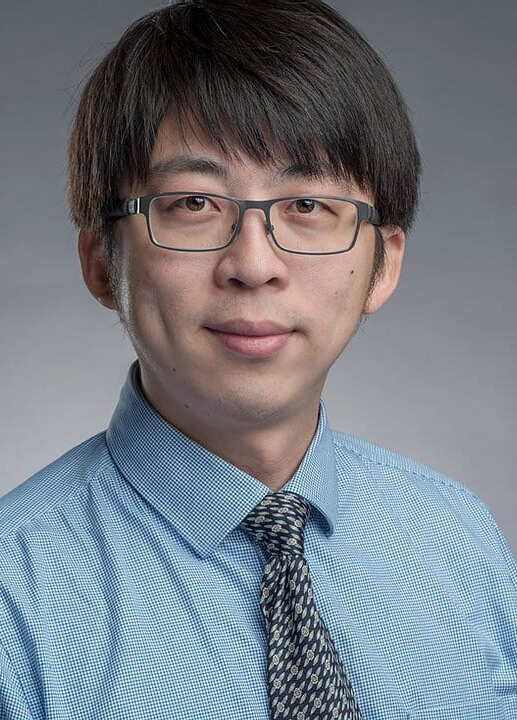}}]{Jiaqi Ma}
(Senior Member, IEEE) is the Director of the FHWA/UCLA Center of Excellence on New Mobility and Automated Vehicles, a Professor at the UCLA Samueli School of Engineering, Director of the UCLA Mobility Lab, and Associate Director of the UCLA Institute of Transportation Studies. He has led and managed numerous research projects funded by the U.S. Department of Transportation, National Science Foundation, state Departments of Transportation, and other federal, state, and local agencies. His primary research interests include automated driving, mobile robotics, multimodal sensing, cooperative perception and decision-making, mobility digital twins, spatial data mining, reasoning, and simulation. Dr. Ma serves as Editor-in-Chief of the IEEE Open Journal of Intelligent Transportation Systems. He is Chair of the Transportation Research Board (TRB) Standing Committee on Connected and Automated Vehicle Systems and a member of the Board of Governors of the IEEE Intelligent Transportation Systems Society.
\end{IEEEbiography}

\end{document}